\documentclass[sigconf]{acmart}

\AtBeginDocument{%
  \providecommand\BibTeX{{%
    \normalfont B\kern-0.5em{\scshape i\kern-0.25em b}\kern-0.8em\TeX}}}

\usepackage{subfigure}
\usepackage{colortbl}
\usepackage{soul}  
\usepackage{enumitem}
\usepackage{listings}
\usepackage{xcolor}
\definecolor{skyblue}{rgb}{0., 0.72, 0.92}
\usepackage{amsmath, amsfonts}
\usepackage[linesnumbered,ruled,vlined]{algorithm2e}
\SetKwInOut{Input}{Input}
\SetKwInOut{Output}{Output}

\SetCommentSty{mycommfont}
\DeclareMathOperator*{\argmax}{arg\,max}

\usepackage{multirow}

\definecolor{lavendermist}{rgb}{0.9, 0.9, 0.9}
\newcommand{\finding}[1]{
    \begingroup
    \sethlcolor{lavendermist}
    \textcolor{black}{\hl{#1}}
    \endgroup
}

\definecolor{lightgray}{rgb}{0.78, 0.78, 0.78}
\sethlcolor{lightgray}  
\definecolor{answercolor}{RGB}{240, 240, 240}
\begin{document}


\title{Characterizing and Understanding the Behavior of Quantized Models for Reliable Deployment}

\author{Qiang Hu}
\affiliation{%
  \institution{SnT, University of Luxembourg}
  \country{Luxembourg}}

\author{Yuejun Guo}
\affiliation{%
  \institution{SnT, University of Luxembourg}
  \country{Luxembourg}}

\author{Maxime Cordy}
\affiliation{%
 \institution{SnT, University of Luxembourg}
  \country{Luxembourg}}
  
\author{Xiaofei Xie}
\affiliation{%
 \institution{Singapore Management University}
 \country{Singapore}}

\author{Wei Ma}
\affiliation{%
  \institution{SnT, University of Luxembourg}
  \country{Luxembourg}}

\author{Mike Papadakis}
\affiliation{%
  \institution{SnT, University of Luxembourg}
  \country{Luxembourg}}

\author{Yves Le Traon}
\affiliation{%
  \institution{SnT, University of Luxembourg}
  \country{Luxembourg}}

\renewcommand{\shortauthors}{Hu and Guo, et al.}

\begin{abstract}
Deep Neural Networks (DNNs) have gained considerable attention in the past decades due to their astounding performance in different applications, such as natural language modeling, self-driving assistance, and source code understanding. With rapid exploration, more and more complex DNN architectures have been proposed along with huge pre-trained model parameters. The common way to use such DNN models in user-friendly devices (e.g., mobile phones) is to perform model compression before deployment. However, recent research has demonstrated that model compression, e.g., model quantization, yields accuracy degradation as well as outputs disagreements when tested on unseen data. Since the unseen data always include distribution shifts and often appear in the wild, the quality and reliability of quantized models are not ensured. In this paper, we conduct a comprehensive study to characterize and help users understand the behaviors of quantized models. Our study considers 4 datasets spanning from image to text, 8 DNN architectures including feed-forward neural networks and recurrent neural networks, and 42 shifted sets with both synthetic and natural distribution shifts. The results reveal that 1) data with distribution shifts happen more disagreements than without. 2) Quantization-aware training can produce more stable models than standard, adversarial, and Mixup training. 3) Disagreements often have closer top-1 and top-2 output probabilities, and $Margin$ is a better indicator than the other uncertainty metrics to distinguish disagreements. 4) Retraining with disagreements has limited efficiency in removing disagreements. We opensource our code and models as a new benchmark for further studying the quantized models.

\end{abstract}



\maketitle

\section{Introduction}

Thanks to the massively available data released and powerful hardware devices supported, Deep Learning (DL) gains considerable attention and achieves even better performance than humans on some tasks \cite{silver2016mastering}. As the backbone of DL systems, Deep Neural Networks (DNNs) follow the data-driven paradigm to learn knowledge from the labeled data automatically and make predictions for incoming unlabelled ones. Inspired by the usage of DNNs for natural language processing, researchers also employ DNNs for source code-related tasks, e.g., code summarization \cite{alon2019code2vec} and problem classification \cite{puri2021project}. Correspondingly, the behavior, quality, and security of DNNs are also concerned by the software engineering community.

A factor that limits the application of DNNs is that DNNs are large and require strong computing resources. For example, the famous language prediction model GPT-3 \cite{brown2020language} has 175 billion parameters, which is hard to be deployed in our daily used devices. For code tasks, the recently released model GraphCodeBERT \cite{guo2020graphcodebert} occupies 124M of storage memory, which is also difficult to be plugged in the generally used IDEs. Furthermore, with the rapid research progress, more and more complex DNNs are developed, which makes the DNN deployment even more challenging.

To solve this deployment issue, instead of directly transferring DNNs to devices, one typical process is to reduce the size of DNN models by model compression for lighter and easier deployment. There are different ways to perform model compression, e.g., model pruning which removes useless parameters from the model, and model quantization which degrades float-level parameters to lower-level parameters (integer-level). In general, the compression process is of great importance and should preserve the performance of original models as much as possible. The reason is that after a model is compressed, it is hard to change it when unexpected problems occur. For example, retraining a model deployed on a mobile device is impractical because this model is packaged.

Unfortunately, recent research has revealed two problems of model compression. First, \cite{guo2019empirical} shows that a compressed model could have a big accuracy difference (more than 5\%) compared to its original model. Second \cite{xie2019diffchaser, tian2021fast} demonstrate that it is common to find inputs that trigger different predictions by a compressed model and its original model. As these study reveal, it remains unclear to what extent model compression preserves prediction performance and under which conditions. The existing literature currently lacks a detailed assessment of these conditions and this lack, in turn, impedes the reliable application of compression techniques.

In this paper, we fill this gap and empirically characterize the behavior of compressed models under various experimental settings in order to better understand the limitations of compression techniques. We specifically consider quantization as this approach is mostly applied in practice \cite{chen2020comprehensive}. We focus our study on the DL models quantized by TensorFLowLite \cite{abadi2016tensorflow} and CoreML \cite{10.5555/3350821} which are widely adopted in the industry. For example, Google uses TensorFLowLite for model deployment on Android devices and Apple applies CoreML for IOS devices. In total, our experimental settings include 4 datasets ranging from image to text, 8 different DNNs including both Feed-forward Neural Networks (FNNs) and Recurrent Neural Networks (RNNs), 42 different sets with both synthetic and natural distribution shifts. With this material, we explore four research questions that existing studies have overlooked:

\textbf{RQ1: How do quantized models react to distribution shifts?} Real applications of DL systems often witness data distributions shifts -- changes in data distribution that typically cause drops in model performance \cite{koh2021wilds}. Given the practical predominance of this phenomenon, research \cite{berend2020cats, hu2021understanding, dola2021distribution} has emphasized the need to consider distribution shift when evaluating DL models. We, therefore, study the impact of model quantization in the case of distribution shifts. We evaluate the quantized models against two types of distribution shift datasets: synthetic (based on image transformations) and natural (reported in the literature). We compare the original and the quantized models in terms of accuracy difference and predicted label differences, i.e. disagreements. 

\textbf{RQ2: How does the training strategy influence the behavior of quantized models?} We explore the influence of different training strategies: standard training which is the basic way to prepare pretrained model, quantization-aware training \cite{jacob2018quantization} which is specifically designed for model quantization, adversarial training \cite{goodfellow2014explaining} and mixup training \cite{zhang2017mixup}, which are the commonly used data augmentation training strategies. We apply each strategy to train original models and then quantize these models. We compare the pairs of models in terms of accuracy difference and disagreements.

\textbf{RQ3: What are the characteristics of the data on which original and quantized models disagree?} We aim to find discriminating factors that can help identify the disagreement inputs. In particular, we investigate whether the most uncertain data are the most likely to produce disagreements. Based on different uncertainty metrics, we train simple classifiers based on logistic regression and evaluate their capabilities to predict disagreements.

\textbf{RQ4: Can model retraining reduce disagreements?} We investigate whether retraining -- a common approach to improve DL models -- can efficiently fix disagreements. Specifically, we explore whether retraining the original model (for additional epochs) with disagreement inputs can help preserve the knowledge of these inputs through the quantization process, and make the quantized model classify these inputs correctly.

In summary, the main novel contributions of this paper are:

\begin{itemize}[leftmargin=*]
    \item We show that synthetic distribution shift has a significant impact on quantized models; it increases the accuracy change by up to 3.03\% and the percentage of disagreements by 5.28\%. 
    
    
    \item We empirically confirm that quantization-aware training is the best method to alleviate performance loss and disagreements after quantization, including when distribution shifts occur.
    
    \item We demonstrate that data uncertainty -- as captured by the Margin metric -- is a suitable factor to discriminate disagreement data. A simple classifier based on Margin reaches an AUC-ROC of 0.63 to 0.97. 
    
    \item We illustrate that retraining on disagreement inputs does not decrease the total level of disagreements between original and quantized models, because it has the side effect of introducing new disagreements.
    
\end{itemize}

We release all our code, models (before and after quantization) and benchmarks in order to support future research studying and improving model quantization\footnote{\url{https://github.com/Anony4paper/quan_study}\label{site}}.


\section{Background}
\label{sec: background}

\subsection{Deep Learning}
\label{subsec:dl}
Deep learning \cite{goodfellow2016deep} is a machine learning technique that uses intermediate layers to progressively obtain knowledge from raw data, and deep neural networks form the backbone of deep learning. A typical deep neural network consists of an input layer, several hidden layers, and an output layer. Each layer includes neurons that mimic the neurons in human brains and undertake specific computations, such as sigmoid and rectifier. The connections between successive layers establish the data flow. In brief, training a deep neural network is to tune the parameters (importance of neurons) of the connections, and testing is to ensure accuracy and reliability during deployment in real-world applications. 

\subsection{Model Quantization}
\label{subsec:mq}
Model quantization is an optimization technique that aims at transforming the higher-bit level weights to lower-bit level weights, e.g., from float32 weights to 8-bit integer weights, to reduce the size of the model for an easy model deployment. Multiple quantization approaches \cite{shomron2021post, pmlr-v139-hubara21a, li2021brecq, jacob2018quantization} have been proposed given its importance in DL-based engineering. An important part of quantization methods is the mapping between the two parts of weights. This mapping can be constructed by using a simple linear function to find the scale for two levels of weights, or by different clustering metrics (e.g., k-means cluster used in CoreML) to find the lookup table quantization of weights.

\begin{figure}[h]
	\centering
	\includegraphics[width=0.3\textwidth]{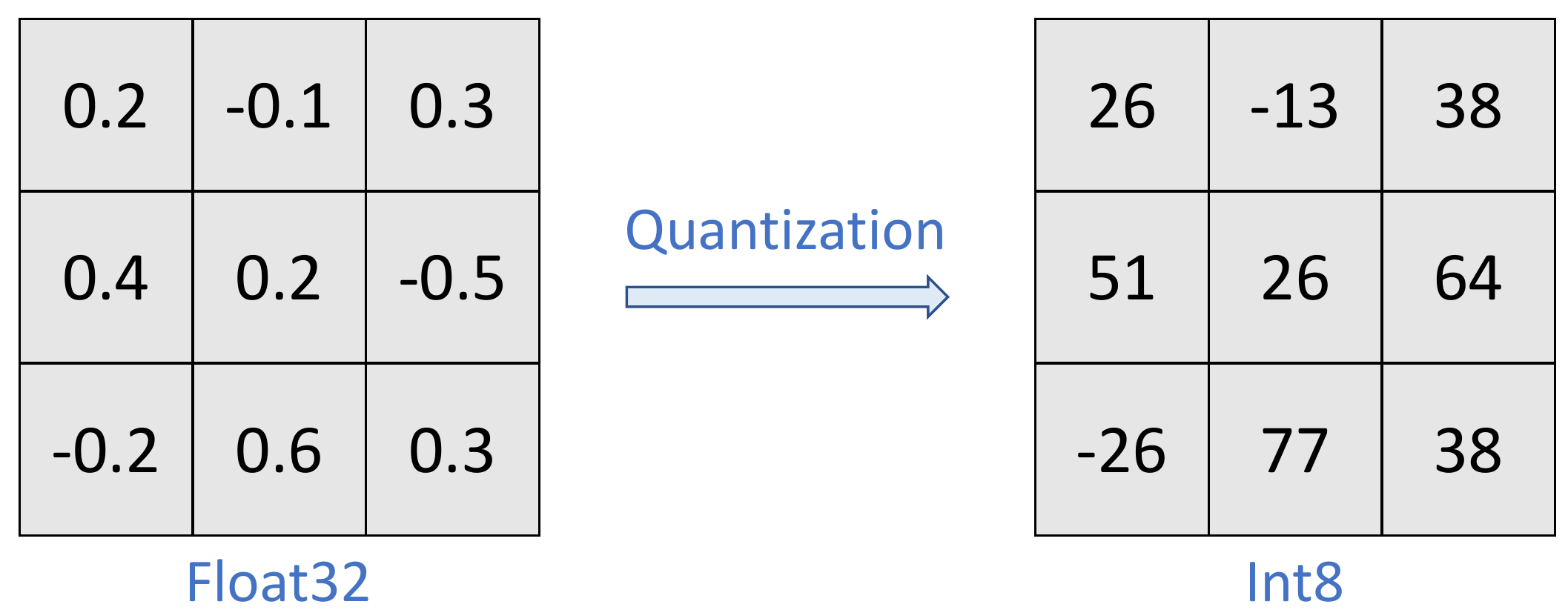}
	\caption{An example of weights quantization. Each weight in Float32 format is converted into Int8.}
	\label{fig:quantization}
\end{figure}

Figure \ref{fig:quantization} gives an example of the basic linear quantization. We assume the left side is the float32-level weights, and the range of these weights is [-1, 1]. We plan to convert the float weights to 8-bit integer weights ranging in [-128, 127]. Thus, the $scale$ here is 128 and the quantized weights is calculated by $Round(weights_{float} / scale)$. Generally, to further reduce the memory usage of quantized models, the quantization only keeps positive weights.

\subsection{Distribution Shift}
\label{subsec:ds}
Distribution shift refers to the change of data distribution in the test dataset compared to the training dataset. Generally, benchmark datasets \cite{lecun1998gradient, maas2011learning} are designed to include training and test data following the same distribution. However, in real-world deployments, the test data can be from the same or a different distribution, which raises the security concern \cite{berend2020cats}. Generally speaking, there are two types of distribution shifts, synthetic and natural \cite{koh2021wilds}. 

Synthetic distribution shift considers possible perturbations in the real world. In addition, concerning the severity of corruption, data can have various levels of noise, which covers many different situations. As a result, synthetic distribution shift is always taken as a starting point to evaluate the performance of a DNN under different settings. A wide range of visual corruptions has been developed in the image domain \cite{mu2019mnist, hendrycks2019benchmarking}. For example, adding motion blur into an image can mimic the scenario of a moving object, and inserting the fog effect can simulate the condition of foggy weather. 

Differ from synthetic shift, natural distribution shift comes from natural variations in datasets. For instance, in the widely used text dataset, IMDb \cite{maas2011learning}, data (movies reviews) are collected from IMDb. When testing, the reviews can be from another movie review website or from a different customer groups.

\section{Overview}
\label{sec: overview}

\subsection{Study Design}

\begin{figure*}[h]
	\centering
	\includegraphics[width=.9\textwidth]{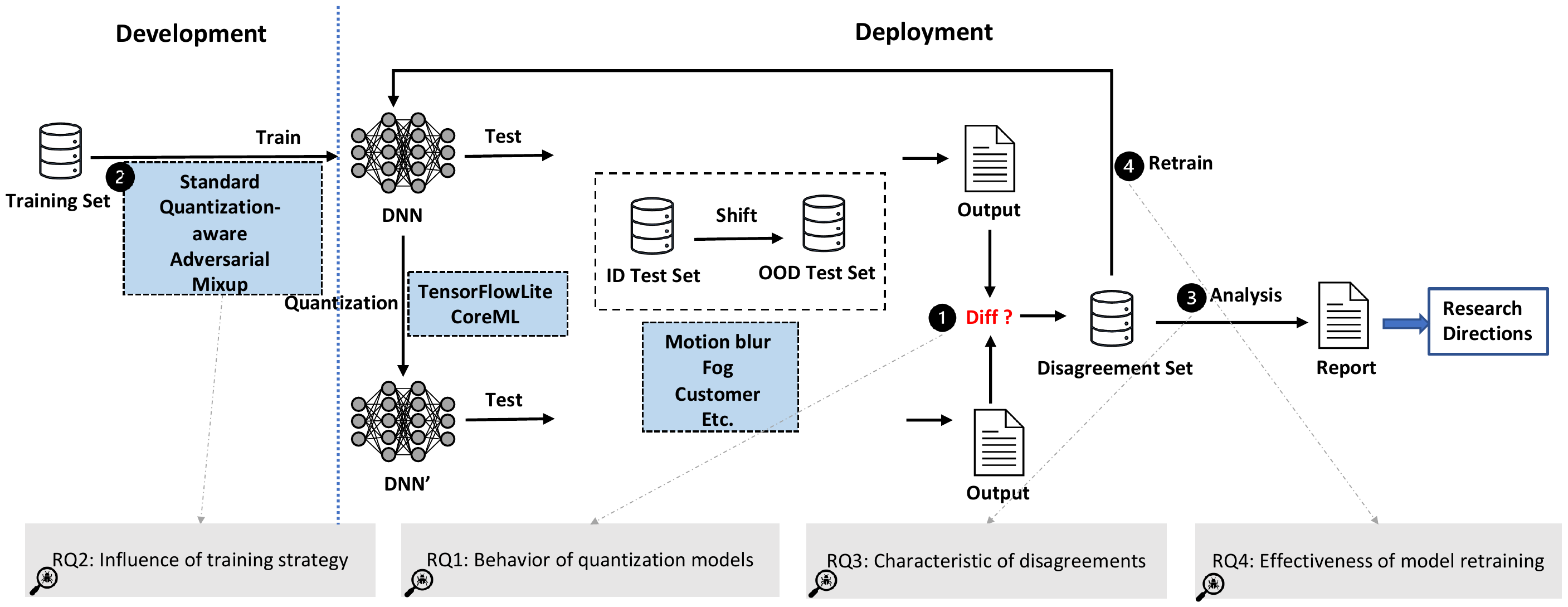}
	\caption{Overview of the experimental design.}
	\label{fig:overview}
\end{figure*}

Figure \ref{fig:overview} gives an overview of our study. Following the common DL systems development process, we prepare the original model \emph{DNN} by standard model training (Section \ref{subsec:ts}) using the collected datasets. Then, we use quantization techniques (e.g., TensorflowLite, CoreML) to compress the model and prepare the optimized model \textit{DNN'} for further deployment. Afterward, to study whether the quantization is reliable or not,  we prepare two types of test data, the ID test set and OOD test set. Remark that the ID test set is the original test data from each dataset, which is in distribution compared to the training data. The OOD test set is the data with distribution shifts. We compare the performance of the original \textit{DNN} and compressed \textit{DNN'} on these two types of test sets and check the \textit{differences} to answer \textbf{RQ1}. In our study, we consider two types of distribution shifts, synthetic and natural.


In the development phase, in addition to standard training, other training strategies can be used to prepare pre-trained models. Thus, it is essential to explore the potential factor that could influence the behaviors of quantized models -- training strategy. We utilize 3 additional training strategies to train the original models and then analyze the behaviors of the quantized model to answer \textbf{RQ2}. Specifically, we include quantization-aware training \cite{jacob2018quantization}, which is specifically designed for solving the problem of accuracy decline after quantization, adversarial training \cite{goodfellow2014explaining} and Mixup training \cite{zhang2017mixup} which aim to improve the generalization of a DNN model.

After analyzing the behaviors of quantized models, we obtain multiple models that are waiting for repair with their disagreements. Before trying to remove the disagreements and repair the quantized models,  the first step should be to investigate the properties of the data that cause disagreements between \emph{DNN} and \emph{DNN'}. We utilize the uncertainty metric as an indicator to check if it can represent the properties of disagreements and answer \textbf{RQ3}. Specifically, for each test dataset (ID/OOD) and model, we collect all the disagreements that have at least once been predicted differently by the original model and the quantized model. Then, we randomly select the same number as the disagreements of normal inputs where the predictions before and after quantization are consistent. Afterward, we obtain the output probabilities of these two (disagreements and normal inputs) sets and calculate their uncertainty scores by different uncertainty metrics as the input data of the logistic regression classifier. We assign the label of disagreement and normal input as 1 and 0, respectively. We then combine and shuffle the two sets and split them into training data and test data following the ratio 9:1. Finally, we train the classifier using the training data and calculate the AUC-ROC score of the classifiers using the prediction of test data with a threshold of 95\%. The AUC-ROC score is used to determine the best uncertainty metric that is discriminative between disagreements and normal inputs significantly.

Finally, we make the first step to repair the quantized model for reliable model deployment. We verify if model retraining is helping to alleviate disagreements to answer \textbf{RQ4}. Model retraining is the most straightforward and commonly used method during deployment to specifically let a pre-trained model work on unlearnt features \cite{hu2022dat}. However, its effectiveness on model quantization is uncovered. After retraining, we follow the same procedure as RQ1 to produce the quantized model and check if the disagreements decreased. Remarkably, we consider both the existing and newly generated disagreements.

\subsection{Datasets and Models}

Table \ref{tab:data_model} presents the details of datasets and models. In this study, we consider 4 widely studied datasets over image and text domains. For each dataset, we build two different models. More specifically, MNIST \cite{lecun1998gradient} is a gray-scale image dataset containing digit numbers from 0 to 9. We train LeNet-1 and LeNet-5 from the LeNet \cite{lecun1998gradient} family. CIFAR-10 contains color images of airplanes and birds. For this dataset, we build two models, Network in network (NiN) \cite{lin2013network} and ResNet-20 \cite{7780459}. iWildCam is a dataset from the distribution shift benchmark Wilds \cite{koh2021wilds}. It consists of color images of different animals, e.g., cow, wild horse, and giraffe. We follow the recommendation of the benchmark to build ResNet-50 \cite{7780459} for iWildCam and add one more model, DenseNet-121 \cite{huang2017densely}, in our study. IMDb \cite{maas2011learning} is a text dataset collected from the popular movie review website IMDb. This dataset is mainly used for sentiment analysis, i.e., the reviewer holds a positive or negative opinion in a movie. We build two well-known RNN models, LSTM \cite{hochreiter1997long} and GRU \cite{chung2014empirical}, for IMDb.  

\begin{table}[!t]
\caption{Details of datasets and DNNs}
\label{tab:data_model}
\resizebox{\columnwidth}{!}{
\begin{tabular}{ccccccc}
\hline
\textbf{Dataset} & \textbf{DNN} & \textbf{Classes} & \textbf{Training} & \textbf{ID Test} & \textbf{Accuracy (\%)} & \textbf{OOD Test} \\ \hline
\cellcolor[HTML]{FFFFFF} & LeNet-1 & \cellcolor[HTML]{FFFFFF} & \cellcolor[HTML]{FFFFFF} & \cellcolor[HTML]{FFFFFF} & 98.62 & \cellcolor[HTML]{FFFFFF} \\
\multirow{-2}{*}{\cellcolor[HTML]{FFFFFF}\textbf{MNIST}} & LeNet-5 & \multirow{-2}{*}{\cellcolor[HTML]{FFFFFF}10} & \multirow{-2}{*}{\cellcolor[HTML]{FFFFFF}60000} & \multirow{-2}{*}{\cellcolor[HTML]{FFFFFF}10000} & 98.87 & \multirow{-2}{*}{\cellcolor[HTML]{FFFFFF}\begin{tabular}[c]{@{}c@{}}MNIST-C\\ (Synthetic)\end{tabular}} \\ \hline
\cellcolor[HTML]{FFFFFF} & ResNet20 & \cellcolor[HTML]{FFFFFF} & \cellcolor[HTML]{FFFFFF} & \cellcolor[HTML]{FFFFFF} & 87.44 & \cellcolor[HTML]{FFFFFF} \\
\multirow{-2}{*}{\cellcolor[HTML]{FFFFFF}\textbf{CIFAR-10}} & NiN & \multirow{-2}{*}{\cellcolor[HTML]{FFFFFF}10} & \multirow{-2}{*}{\cellcolor[HTML]{FFFFFF}50000} & \multirow{-2}{*}{\cellcolor[HTML]{FFFFFF}10000} & 88.27 & \multirow{-2}{*}{\cellcolor[HTML]{FFFFFF}\begin{tabular}[c]{@{}c@{}}CIFAR-10-C\\ (Synthetic)\end{tabular}} \\ \hline
\cellcolor[HTML]{FFFFFF} & ResNet-50 & \cellcolor[HTML]{FFFFFF} & \cellcolor[HTML]{FFFFFF} & \cellcolor[HTML]{FFFFFF} & 75.78 & \cellcolor[HTML]{FFFFFF} \\
\multirow{-2}{*}{\cellcolor[HTML]{FFFFFF}\textbf{iWildCam}} & Densenet-121 & \multirow{-2}{*}{\cellcolor[HTML]{FFFFFF}182} & \multirow{-2}{*}{\cellcolor[HTML]{FFFFFF}129809} & \multirow{-2}{*}{\cellcolor[HTML]{FFFFFF}8154} & 76.01 & \multirow{-2}{*}{\cellcolor[HTML]{FFFFFF}\begin{tabular}[c]{@{}c@{}}Camera Traps\\ (Natural)\end{tabular}} \\ \hline
\cellcolor[HTML]{FFFFFF} & LSTM & \cellcolor[HTML]{FFFFFF} & \cellcolor[HTML]{FFFFFF} & \cellcolor[HTML]{FFFFFF} & 83.78 & \cellcolor[HTML]{FFFFFF} \\
\multirow{-2}{*}{\cellcolor[HTML]{FFFFFF}\textbf{IMDb}} & GRU & \multirow{-2}{*}{\cellcolor[HTML]{FFFFFF}2} & \multirow{-2}{*}{\cellcolor[HTML]{FFFFFF}5000} & \multirow{-2}{*}{\cellcolor[HTML]{FFFFFF}5000} & 83.14 & \multirow{-2}{*}{\cellcolor[HTML]{FFFFFF}\begin{tabular}[c]{@{}c@{}}CR, Yelp\\ (Natural)\end{tabular}} \\ \hline

\end{tabular}
}
\end{table}

\textbf{Test data with distribution shift.} For synthetic distribution shift, we test on MNIST and CIFAR-10 with benchmark datasets MNIST-C \cite{mu2019mnist} and CIFAR-10-C \cite{hendrycks2019benchmarking}, respectively. Both benchmarks include several groups of noisy images synthesized by different image transformation methods, e.g., image rotation and image scale. MNIST-C contains 16 types of transformations and CIFAR-10-C has 19 types. For natural distribution shift,  we test on iWildCam and IMDb using the Wilds benchmark. The distribution shift comes from the change of camera traps in iWildCam and the difference in websites and customers in IMDb.

\subsection{Quantization Techniques}

\textbf{TensorflowLite} \cite{abadi2016tensorflow} is a component of the deep learning framework -- TensorFlow, which is developed and maintained by Google. It provides interfaces to covert TensorFlow models into Lite models to promote the deployment in different low-computing devices, such as  Android mobile phones. Currently, TensorFLowLite supports both 8-bit integer and 16-bit float quantizations for most DNNs except 8-bit integer quantization for RNNs \cite{issue35194}. In our experiments, we only apply 16-bit float quantization for IMDb-related models.

\textbf{CoreML} \cite{10.5555/3350821} is an Apple framework that converts models from third-party frameworks (e.g., TensorFlow and Pytorch) to Mlmodel. Mlmodel is a specific deep learning model format for IOS platforms. CoreML also provides post-training quantization interfaces to compress models. Differ from TensorflowLite, CoreML supports all bits level quantization for all types of DNNs.

\subsection{Training Strategies}
\label{subsec:ts}
In addition to standard training, we consider three representative training strategies from different perspectives, quantization-aware \cite{jacob2018quantization}, adversarial \cite{goodfellow2014explaining}, and Mixup \cite{zhang2017mixup}.

\textbf{Standard training} is the baseline to evaluate the other training strategies. In this setting, we train the model without any modification in the model (e.g., quantization-aware) or data (e.g., Mixup).

\textbf{Quantization-aware training} is designed by the TensorFlow group, which is used for preserving the accuracy of the model after post-training quantization in the training process. It simulates the quantization effects in the forward pass of training. Namely, during training, the parameters of the model will be updated by both the normal operations and the injected quantization operations. In this way, the trained model can learn the knowledge for quantization.

\textbf{Adversarial training} is one of the most effective defenses for promoting model robustness by adversarially data augmentation. Compared to standard training, adversarial examples crafted from raw inputs are fed to train the model during each epoch. As a result, the training dataset is augmented successively. 

\textbf{Mixup training} is a data augmentation technique that generates new samples by weighted combinations of random training data and their labels. It has been empirically proved to be effective in improving the generalization of DNNs and has several variants, such as AugMix \cite{hendrycks2019augmix}. In this paper, we consider the original Mixup.

\subsection{Evaluation Measures}
We consider both the \textbf{accuracy} and \textbf{disagreement} to evaluate the performance of DNNs, and use \textbf{AUC-ROC} to evaluate the performance of logistic regression classifiers.

\textbf{Accuracy} is the basic criterion to quantify the quality of a DNN model, which refers to the ratio of correct predictions. 

\textbf{Number of disagreements} is defined in \cite{xie2019diffchaser} to characterize the difference between two DNNs. A disagreement is an input that triggers different outputs by the original model and its quantized version. By measuring the number of disagreements in the test data, one can observe the model's behavior change after quantization.

\textbf{Area Under the Receiver Operating Characteristic Curve (AUC-ROC)} \cite{fawcett2006introduction} is a threshold-independent performance evaluation metric. In RQ3, we utilize AUC-ROC score to measure the performance of the trained logistic regression classifiers.



\subsection{Uncertainty Metrics}
In RQ3, we utilize uncertainty metrics to estimate the characteristics of the disagreement inputs. Following previous studies \cite{hu2021towards, ma2021test}, we select 4 commonly used output-based uncertainty metrics in our study. Given a classification task, let $DNN$ be a $C$-class model and $x$ be an input. $p_i\left(x\right)$ denotes the predicted probability of $x$ belonging to the $i$th class, $0 \leq i \leq C$. \textbf{Entropy} score \cite{shannon1948mathematical} quantifies the uncertainty of $x$ by Shannon entropy: $Entropy(x)$ = -$\sum_{i=1}^{C}p_i\left(x\right)\log p_i\left(x\right)$. \textbf{Gini} \cite{feng2020deepgini} score is calculated as: $Gini(x)$ = $1-\sum_{i=1}^{C}\left(p_i\left(x\right)\right)^2$. $Margin$ \cite{wang2014new} score is based on the top-2 prediction probabilities: $Margin(x)$ = $Margin\left(x\right)=p_k\left(x\right)-p_j\left(x\right)$, where $k=\underset{i=1:C}{\argmax}\left(p_i\left(x\right)\right)$ and $j=\underset{{i=\{1:C\}/k}}{\argmax}\left(p_i\left(x\right)\right)$. \textbf{Least Confidence (LC)} \cite{settles2009active} score is the difference between the most confident prediction and 100\% confidence. $LC(x)$ = 1 - $p_k\left(x\right)$, where $k=\underset{i=1:C}{\argmax}\left(p_i\left(x\right)\right)$.









\section{Configuration}
\label{sec: implementation}

\textbf{Environments.} We undertake model training and retraining on an NVIDIA Tesla V100 16G SXM2 GPU. For the TensorFlowLite model evaluation, we run experiments on a 2.6 GHz Intel Xeon Gold 6132 CPU. For the CoreML model evaluation, we conduct experiments on a MacBook Pro laptop with macOS Big Sur 11.0.1 with a 2GHz GHz QuadCore Intel Core i5 CPU with 16GB RAM.

\textbf{Quantization.} We apply the interfaces provided by TensorFLowLite and CoreML to accomplish post-training model quantization. For IMDb-related models, we only apply 16-bit float quantization by TensorFlowLite and utilize both 8-bit interger and 16-bit float quantization by CoreML. For other models, we conduct 8-bit integer and 16-bit float quantization using both techniques.

\textbf{Model training.} For the quantization-aware training, we mask layers (e.g., BatchNormalization layer) that are not supported by the current TensorFlow framework. In addition, since TensorFlow does not support RNNs \cite{issue25563}, we skip IMDb-related models in this experiment. Regarding the adversarial training, we employ the commonly used PGD-based \cite{madry2017towards} adversarial training for image datasets, and PWWS-based \cite{ren2019generating} adversarial training for text datasets. Concerning the Mixup training, we follow the recommendation by the original paper to set the mixup parameter $\alpha$ as 0.2. 

\textbf{Model retraining.} Following the same setting from the empirical study of model retraining \cite{hu2022dat}, we add all disagreements into original training data to train the pre-trained model with additional several epochs (5 epochs for MNIST, IMDb, and iWildsCam, 10 epochs for CIFAR-10). 

All the detailed configurations can be found at our project site \ref{site}.

\section{Experimental Results}
\label{sec: results}
In this section, we report the experimental results to answer each research question. Meanwhile, we highlight our novel findings.

\subsection{RQ1: Behavior of Quantized Models}
\label{subsec:rq1}
Table \ref{tab:rq1_sds} presents the results of the behaviors of quantized models on ID test data and OOD test data with synthetic distribution shifts. Concerning the accuracy change, the accuracy is supposed to degrade due to the loss of information during quantization, which is also demonstrated by the existing studies \cite{guo2019empirical, hu2021towards}. Surprisingly, the results also show almost 30\% of (86 out of 292) opposite cases where quantized models hold higher accuracy than their original models. Particularly, in the case of $ResNet20, Zoom\_blur$, the quantized model has an improvement of 1.98\%. On the other hand, this phenomenon also happens to the natural distribution shift (10 out of 16 cases in Table \ref{tab:rq1_nds}). Regarding shifted data as natural adversarial examples, our finding confirms the conclusion from a recent research \cite{pmlr-v139-fu21c} that the quantization process can be useful to promote the model's adversarial robustness. In addition, the distribution shift can lead to larger change and should be taken into account during deployment. For example, in \emph{MNIST, TF-8}, the quantized model has an accuracy change of 0.04\% on ID test data but 0.78 under the Fog shift (Table \ref{tab:rq1_sds}). And comparing the ID and OOD test sets, we found the synthetic distribution shift can increase the accuracy change by up to 3.03\% (ResNet20-Gaussian\_noise-CM-8). \finding{\textbf{Finding 1}: Post-training model quantization does not always harm the accuracy of original models. Distribution shift can cause a large accuracy change compared to testing on ID data.}

\begin{table}
\caption{Behavior of quantized models under synthetic distribution shift. Non-highlighted value: accuracy change (\%), \hl{highlighted value}: number of disagreements. A low value indicates a small difference between the original and quantized models. ID refers to the ID test data and the others are OOD test data. TF: TensorFlowLite. CM: CoreML. |Average|: the average of absolute changes.}
\label{tab:rq1_sds}
\centering
\resizebox{\columnwidth}{!}{
\begin{tabular}{lcccccccccccccccc}
\hline
 & \multicolumn{16}{c}{\textbf{MNIST}} \\ \cline{2-17} 
 & \multicolumn{8}{c}{\textbf{LeNet1}} & \multicolumn{8}{c}{\textbf{LeNet5}} \\
\multirow{-3}{*}{\textbf{Test Data}} & \multicolumn{2}{c}{\textbf{TF-8}} & \multicolumn{2}{c}{\textbf{TF-16}} & \multicolumn{2}{c}{\textbf{CM-8}} & \multicolumn{2}{c}{\textbf{CM-16}} & \multicolumn{2}{c}{\textbf{TF-8}} & \multicolumn{2}{c}{\textbf{TF-16}} & \multicolumn{2}{c}{\textbf{CM-8}} & \multicolumn{2}{c}{\textbf{CM-16}} \\ \hline
\textbf{ID} & -0.04 & \cellcolor[HTML]{C0C0C0}14 & -0.04 & \cellcolor[HTML]{C0C0C0}9 & 0 & \cellcolor[HTML]{C0C0C0}2 & 0 & \multicolumn{1}{c|}{\cellcolor[HTML]{C0C0C0}0} & 0.02 & \cellcolor[HTML]{C0C0C0}4 & 0.01 & \cellcolor[HTML]{C0C0C0}3 & 0.01 & \cellcolor[HTML]{C0C0C0}1 & 0 & \cellcolor[HTML]{C0C0C0}0 \\
\textbf{Brightness} & 0.51 & \cellcolor[HTML]{C0C0C0}172 & 0.28 & \cellcolor[HTML]{C0C0C0}67 & -0.04 & \cellcolor[HTML]{C0C0C0}53 & 0.01 & \multicolumn{1}{c|}{\cellcolor[HTML]{C0C0C0}7} & -0.27 & \cellcolor[HTML]{C0C0C0}175 & -0.56 & \cellcolor[HTML]{C0C0C0}100 & -0.79 & \cellcolor[HTML]{C0C0C0}100 & 0.03 & \cellcolor[HTML]{C0C0C0}6 \\
\textbf{Canny\_edges} & 0.77 & \cellcolor[HTML]{C0C0C0}172 & 0.5 & \cellcolor[HTML]{C0C0C0}86 & 0.02 & \cellcolor[HTML]{C0C0C0}51 & -0.01 & \multicolumn{1}{c|}{\cellcolor[HTML]{C0C0C0}4} & 0.16 & \cellcolor[HTML]{C0C0C0}73 & -0.06 & \cellcolor[HTML]{C0C0C0}35 & -0.01 & \cellcolor[HTML]{C0C0C0}17 & 0.02 & \cellcolor[HTML]{C0C0C0}2 \\
\textbf{Dotted\_line} & -0.21 & \cellcolor[HTML]{C0C0C0}38 & -0.06 & \cellcolor[HTML]{C0C0C0}24 & -0.03 & \cellcolor[HTML]{C0C0C0}8 & 0 & \multicolumn{1}{c|}{\cellcolor[HTML]{C0C0C0}0} & -0.01 & \cellcolor[HTML]{C0C0C0}26 & -0.06 & \cellcolor[HTML]{C0C0C0}13 & -0.04 & \cellcolor[HTML]{C0C0C0}12 & 0 & \cellcolor[HTML]{C0C0C0}0 \\
\textbf{Fog} & 0.78 & \cellcolor[HTML]{C0C0C0}542 & 0.11 & \cellcolor[HTML]{C0C0C0}133 & -0.17 & \cellcolor[HTML]{C0C0C0}112 & 0 & \multicolumn{1}{c|}{\cellcolor[HTML]{C0C0C0}6} & 0.31 & \cellcolor[HTML]{C0C0C0}321 & -0.43 & \cellcolor[HTML]{C0C0C0}112 & -0.6 & \cellcolor[HTML]{C0C0C0}120 & 0.01 & \cellcolor[HTML]{C0C0C0}8 \\
\textbf{Glass\_blur} & -0.05 & \cellcolor[HTML]{C0C0C0}41 & -0.04 & \cellcolor[HTML]{C0C0C0}18 & -0.05 & \cellcolor[HTML]{C0C0C0}10 & 0 & \multicolumn{1}{c|}{\cellcolor[HTML]{C0C0C0}0} & 0.09 & \cellcolor[HTML]{C0C0C0}44 & -0.1 & \cellcolor[HTML]{C0C0C0}23 & -0.03 & \cellcolor[HTML]{C0C0C0}17 & 0 & \cellcolor[HTML]{C0C0C0}0 \\
\textbf{Identity} & -0.04 & \cellcolor[HTML]{C0C0C0}14 & -0.04 & \cellcolor[HTML]{C0C0C0}9 & 0 & \cellcolor[HTML]{C0C0C0}2 & 0 & \multicolumn{1}{c|}{\cellcolor[HTML]{C0C0C0}0} & 0.02 & \cellcolor[HTML]{C0C0C0}4 & 0.01 & \cellcolor[HTML]{C0C0C0}3 & 0.01 & \cellcolor[HTML]{C0C0C0}1 & 0 & \cellcolor[HTML]{C0C0C0}0 \\
\textbf{Impulse\_noise} & -0.23 & \cellcolor[HTML]{C0C0C0}77 & 0.06 & \cellcolor[HTML]{C0C0C0}28 & -0.11 & \cellcolor[HTML]{C0C0C0}33 & -0.04 & \multicolumn{1}{c|}{\cellcolor[HTML]{C0C0C0}4} & -0.02 & \cellcolor[HTML]{C0C0C0}50 & -0.13 & \cellcolor[HTML]{C0C0C0}35 & 0.01 & \cellcolor[HTML]{C0C0C0}23 & -0.03 & \cellcolor[HTML]{C0C0C0}3 \\
\textbf{Motion\_blur} & 0.14 & \cellcolor[HTML]{C0C0C0}79 & 0.08 & \cellcolor[HTML]{C0C0C0}29 & -0.07 & \cellcolor[HTML]{C0C0C0}20 & 0.01 & \multicolumn{1}{c|}{\cellcolor[HTML]{C0C0C0}1} & 0.18 & \cellcolor[HTML]{C0C0C0}59 & -0.11 & \cellcolor[HTML]{C0C0C0}23 & -0.01 & \cellcolor[HTML]{C0C0C0}17 & 0.01 & \cellcolor[HTML]{C0C0C0}1 \\
\textbf{Rotate} & -0.07 & \cellcolor[HTML]{C0C0C0}62 & -0.03 & \cellcolor[HTML]{C0C0C0}30 & 0 & \cellcolor[HTML]{C0C0C0}13 & -0.02 & \multicolumn{1}{c|}{\cellcolor[HTML]{C0C0C0}2} & -0.11 & \cellcolor[HTML]{C0C0C0}30 & -0.09 & \cellcolor[HTML]{C0C0C0}18 & -0.05 & \cellcolor[HTML]{C0C0C0}10 & 0 & \cellcolor[HTML]{C0C0C0}0 \\
\textbf{Scale} & -0.28 & \cellcolor[HTML]{C0C0C0}102 & -0.15 & \cellcolor[HTML]{C0C0C0}43 & -0.04 & \cellcolor[HTML]{C0C0C0}29 & 0 & \multicolumn{1}{c|}{\cellcolor[HTML]{C0C0C0}0} & -0.05 & \cellcolor[HTML]{C0C0C0}53 & -0.02 & \cellcolor[HTML]{C0C0C0}22 & -0.02 & \cellcolor[HTML]{C0C0C0}8 & 0 & \cellcolor[HTML]{C0C0C0}0 \\
\textbf{Shear} & 0 & \cellcolor[HTML]{C0C0C0}22 & 0 & \cellcolor[HTML]{C0C0C0}8 & 0.01 & \cellcolor[HTML]{C0C0C0}11 & 0.01 & \multicolumn{1}{c|}{\cellcolor[HTML]{C0C0C0}2} & 0 & \cellcolor[HTML]{C0C0C0}22 & -0.03 & \cellcolor[HTML]{C0C0C0}11 & -0.01 & \cellcolor[HTML]{C0C0C0}5 & 0 & \cellcolor[HTML]{C0C0C0}0 \\
\textbf{Shot\_noise} & -0.06 & \cellcolor[HTML]{C0C0C0}26 & -0.02 & \cellcolor[HTML]{C0C0C0}11 & 0 & \cellcolor[HTML]{C0C0C0}10 & 0 & \multicolumn{1}{c|}{\cellcolor[HTML]{C0C0C0}0} & 0.06 & \cellcolor[HTML]{C0C0C0}16 & -0.01 & \cellcolor[HTML]{C0C0C0}7 & -0.03 & \cellcolor[HTML]{C0C0C0}6 & 0 & \cellcolor[HTML]{C0C0C0}0 \\
\textbf{Spatter} & -0.05 & \cellcolor[HTML]{C0C0C0}31 & 0.06 & \cellcolor[HTML]{C0C0C0}13 & 0.02 & \cellcolor[HTML]{C0C0C0}6 & 0 & \multicolumn{1}{c|}{\cellcolor[HTML]{C0C0C0}0} & 0.04 & \cellcolor[HTML]{C0C0C0}14 & -0.02 & \cellcolor[HTML]{C0C0C0}10 & 0 & \cellcolor[HTML]{C0C0C0}8 & -0.01 & \cellcolor[HTML]{C0C0C0}1 \\
\textbf{Stripe} & -0.42 & \cellcolor[HTML]{C0C0C0}113 & -0.1 & \cellcolor[HTML]{C0C0C0}70 & 0.18 & \cellcolor[HTML]{C0C0C0}103 & -0.04 & \multicolumn{1}{c|}{\cellcolor[HTML]{C0C0C0}6} & -0.03 & \cellcolor[HTML]{C0C0C0}88 & -0.03 & \cellcolor[HTML]{C0C0C0}36 & -0.19 & \cellcolor[HTML]{C0C0C0}53 & 0 & \cellcolor[HTML]{C0C0C0}1 \\
\textbf{Translate} & -0.18 & \cellcolor[HTML]{C0C0C0}159 & -0.08 & \cellcolor[HTML]{C0C0C0}70 & -0.04 & \cellcolor[HTML]{C0C0C0}67 & 0 & \multicolumn{1}{c|}{\cellcolor[HTML]{C0C0C0}4} & 0.15 & \cellcolor[HTML]{C0C0C0}135 & -0.05 & \cellcolor[HTML]{C0C0C0}64 & 0 & \cellcolor[HTML]{C0C0C0}47 & -0.01 & \cellcolor[HTML]{C0C0C0}2 \\
\textbf{Zigzag} & 0.03 & \cellcolor[HTML]{C0C0C0}88 & -0.03 & \cellcolor[HTML]{C0C0C0}41 & -0.01 & \cellcolor[HTML]{C0C0C0}34 & -0.04 & \multicolumn{1}{c|}{\cellcolor[HTML]{C0C0C0}7} & -0.06 & \cellcolor[HTML]{C0C0C0}66 & -0.17 & \cellcolor[HTML]{C0C0C0}34 & -0.08 & \cellcolor[HTML]{C0C0C0}30 & -0.01 & \cellcolor[HTML]{C0C0C0}1 \\ \hline
\textbf{|Average|} & 0.23 & \cellcolor[HTML]{C0C0C0}103 & 0.10 & \cellcolor[HTML]{C0C0C0}41 & 0.05 & \cellcolor[HTML]{C0C0C0}33 & 0.01 & \multicolumn{1}{c|}{\cellcolor[HTML]{C0C0C0}3} & 0.09 & \cellcolor[HTML]{C0C0C0}69 & 0.11 & \cellcolor[HTML]{C0C0C0}32 & 0.11 & \cellcolor[HTML]{C0C0C0}28 & 0.01 & \cellcolor[HTML]{C0C0C0}1 \\ \hline
 & \multicolumn{16}{c}{\textbf{CIFAR-10}} \\ \cline{2-17} 
 & \multicolumn{8}{c}{\textbf{NiN}} & \multicolumn{8}{c}{\textbf{ResNet20}} \\
 & \multicolumn{2}{c}{\textbf{TF-8}} & \multicolumn{2}{c}{\textbf{TF-16}} & \multicolumn{2}{c}{\textbf{CM-8}} & \multicolumn{2}{c}{\textbf{CM-16}} & \multicolumn{2}{c}{\textbf{TF-8}} & \multicolumn{2}{c}{\textbf{TF-16}} & \multicolumn{2}{c}{\textbf{CM-8}} & \multicolumn{2}{c}{\textbf{CM-16}} \\ \hline
\textbf{ID} & -0.95 & \cellcolor[HTML]{C0C0C0}514 & -0.1 & \cellcolor[HTML]{C0C0C0}24 & 0.03 & \cellcolor[HTML]{C0C0C0}45 & 0.02 & \multicolumn{1}{c|}{\cellcolor[HTML]{C0C0C0}7} & 0.04 & \cellcolor[HTML]{C0C0C0}456 & -0.4 & \cellcolor[HTML]{C0C0C0}54 & 0.36 & \cellcolor[HTML]{C0C0C0}181 & 0.03 & \cellcolor[HTML]{C0C0C0}7 \\
\textbf{Brightness} & -0.02 & \cellcolor[HTML]{C0C0C0}70 & -0.01 & \cellcolor[HTML]{C0C0C0}50 & 0.03 & \cellcolor[HTML]{C0C0C0}44 & 0.01 & \multicolumn{1}{c|}{\cellcolor[HTML]{C0C0C0}9} & -0.02 & \cellcolor[HTML]{C0C0C0}190 & 0.1 & \cellcolor[HTML]{C0C0C0}51 & -0.08 & \cellcolor[HTML]{C0C0C0}170 & -0.03 & \cellcolor[HTML]{C0C0C0}5 \\
\textbf{Contrast} & 0.04 & \cellcolor[HTML]{C0C0C0}78 & -0.06 & \cellcolor[HTML]{C0C0C0}42 & -0.06 & \cellcolor[HTML]{C0C0C0}48 & -0.04 & \multicolumn{1}{c|}{\cellcolor[HTML]{C0C0C0}6} & -0.12 & \cellcolor[HTML]{C0C0C0}250 & 0.04 & \cellcolor[HTML]{C0C0C0}49 & 0.02 & \cellcolor[HTML]{C0C0C0}187 & -0.06 & \cellcolor[HTML]{C0C0C0}10 \\
\textbf{Defocus\_blur} & -0.1 & \cellcolor[HTML]{C0C0C0}51 & -0.05 & \cellcolor[HTML]{C0C0C0}34 & -0.06 & \cellcolor[HTML]{C0C0C0}37 & -0.04 & \multicolumn{1}{c|}{\cellcolor[HTML]{C0C0C0}5} & -0.21 & \cellcolor[HTML]{C0C0C0}205 & 0.01 & \cellcolor[HTML]{C0C0C0}53 & 0.02 & \cellcolor[HTML]{C0C0C0}175 & -0.03 & \cellcolor[HTML]{C0C0C0}15 \\
\textbf{Elastic\_transform} & 0.03 & \cellcolor[HTML]{C0C0C0}107 & 0.02 & \cellcolor[HTML]{C0C0C0}65 & 0.06 & \cellcolor[HTML]{C0C0C0}70 & 0.01 & \multicolumn{1}{c|}{\cellcolor[HTML]{C0C0C0}10} & 0.02 & \cellcolor[HTML]{C0C0C0}342 & 0 & \cellcolor[HTML]{C0C0C0}84 & 0.83 & \cellcolor[HTML]{C0C0C0}302 & -0.05 & \cellcolor[HTML]{C0C0C0}18 \\
\textbf{Fog} & -0.02 & \cellcolor[HTML]{C0C0C0}57 & 0.02 & \cellcolor[HTML]{C0C0C0}29 & -0.06 & \cellcolor[HTML]{C0C0C0}37 & -0.03 & \multicolumn{1}{c|}{\cellcolor[HTML]{C0C0C0}6} & -0.07 & \cellcolor[HTML]{C0C0C0}236 & 0.01 & \cellcolor[HTML]{C0C0C0}45 & -0.07 & \cellcolor[HTML]{C0C0C0}212 & 0 & \cellcolor[HTML]{C0C0C0}16 \\
\textbf{Frost} & 0.02 & \cellcolor[HTML]{C0C0C0}81 & -0.05 & \cellcolor[HTML]{C0C0C0}53 & 0 & \cellcolor[HTML]{C0C0C0}50 & 0 & \multicolumn{1}{c|}{\cellcolor[HTML]{C0C0C0}10} & -0.28 & \cellcolor[HTML]{C0C0C0}260 & 0.05 & \cellcolor[HTML]{C0C0C0}68 & -0.76 & \cellcolor[HTML]{C0C0C0}317 & 0.01 & \cellcolor[HTML]{C0C0C0}14 \\
\textbf{Gaussian\_blur} & -0.05 & \cellcolor[HTML]{C0C0C0}56 & -0.06 & \cellcolor[HTML]{C0C0C0}29 & -0.04 & \cellcolor[HTML]{C0C0C0}36 & -0.02 & \multicolumn{1}{c|}{\cellcolor[HTML]{C0C0C0}5} & -0.18 & \cellcolor[HTML]{C0C0C0}207 & 0.05 & \cellcolor[HTML]{C0C0C0}60 & 0.04 & \cellcolor[HTML]{C0C0C0}161 & 0.03 & \cellcolor[HTML]{C0C0C0}12 \\
\textbf{Gaussian\_noise} & 0.06 & \cellcolor[HTML]{C0C0C0}96 & -0.03 & \cellcolor[HTML]{C0C0C0}57 & 0.06 & \cellcolor[HTML]{C0C0C0}65 & -0.06 & \multicolumn{1}{c|}{\cellcolor[HTML]{C0C0C0}8} & -0.35 & \cellcolor[HTML]{C0C0C0}343 & 0.06 & \cellcolor[HTML]{C0C0C0}106 & -2.67 & \cellcolor[HTML]{C0C0C0}499 & -0.15 & \cellcolor[HTML]{C0C0C0}29 \\
\textbf{Glass\_blur} & -0.36 & \cellcolor[HTML]{C0C0C0}164 & -0.42 & \cellcolor[HTML]{C0C0C0}122 & -0.02 & \cellcolor[HTML]{C0C0C0}90 & -0.08 & \multicolumn{1}{c|}{\cellcolor[HTML]{C0C0C0}20} & -0.09 & \cellcolor[HTML]{C0C0C0}559 & 0.12 & \cellcolor[HTML]{C0C0C0}168 & -1.56 & \cellcolor[HTML]{C0C0C0}754 & -0.14 & \cellcolor[HTML]{C0C0C0}57 \\
\textbf{Impulse\_noise} & -0.28 & \cellcolor[HTML]{C0C0C0}108 & -0.34 & \cellcolor[HTML]{C0C0C0}86 & -0.11 & \cellcolor[HTML]{C0C0C0}82 & -0.05 & \multicolumn{1}{c|}{\cellcolor[HTML]{C0C0C0}18} & -0.12 & \cellcolor[HTML]{C0C0C0}275 & 0.04 & \cellcolor[HTML]{C0C0C0}72 & -1.1 & \cellcolor[HTML]{C0C0C0}329 & -0.01 & \cellcolor[HTML]{C0C0C0}20 \\
\textbf{Jpeg\_compression} & -0.26 & \cellcolor[HTML]{C0C0C0}82 & -0.15 & \cellcolor[HTML]{C0C0C0}55 & -0.21 & \cellcolor[HTML]{C0C0C0}57 & -0.05 & \multicolumn{1}{c|}{\cellcolor[HTML]{C0C0C0}12} & -0.15 & \cellcolor[HTML]{C0C0C0}259 & -0.08 & \cellcolor[HTML]{C0C0C0}75 & -0.94 & \cellcolor[HTML]{C0C0C0}315 & -0.06 & \cellcolor[HTML]{C0C0C0}21 \\
\textbf{Motion\_blur} & 0 & \cellcolor[HTML]{C0C0C0}87 & -0.06 & \cellcolor[HTML]{C0C0C0}43 & 0.14 & \cellcolor[HTML]{C0C0C0}67 & -0.05 & \multicolumn{1}{c|}{\cellcolor[HTML]{C0C0C0}13} & 0.46 & \cellcolor[HTML]{C0C0C0}336 & -0.04 & \cellcolor[HTML]{C0C0C0}83 & 1.69 & \cellcolor[HTML]{C0C0C0}325 & -0.02 & \cellcolor[HTML]{C0C0C0}21 \\
\textbf{Pixelate} & 0.03 & \cellcolor[HTML]{C0C0C0}84 & 0 & \cellcolor[HTML]{C0C0C0}49 & -0.06 & \cellcolor[HTML]{C0C0C0}53 & -0.01 & \multicolumn{1}{c|}{\cellcolor[HTML]{C0C0C0}6} & 0.08 & \cellcolor[HTML]{C0C0C0}251 & 0.05 & \cellcolor[HTML]{C0C0C0}61 & -0.32 & \cellcolor[HTML]{C0C0C0}239 & -0.02 & \cellcolor[HTML]{C0C0C0}10 \\
\textbf{Saturate} & -0.13 & \cellcolor[HTML]{C0C0C0}89 & 0 & \cellcolor[HTML]{C0C0C0}49 & -0.05 & \cellcolor[HTML]{C0C0C0}49 & -0.02 & \multicolumn{1}{c|}{\cellcolor[HTML]{C0C0C0}8} & -0.16 & \cellcolor[HTML]{C0C0C0}262 & -0.02 & \cellcolor[HTML]{C0C0C0}66 & 0.11 & \cellcolor[HTML]{C0C0C0}238 & -0.01 & \cellcolor[HTML]{C0C0C0}22 \\
\textbf{Shot\_noise} & -0.06 & \cellcolor[HTML]{C0C0C0}105 & -0.08 & \cellcolor[HTML]{C0C0C0}60 & -0.09 & \cellcolor[HTML]{C0C0C0}69 & -0.02 & \multicolumn{1}{c|}{\cellcolor[HTML]{C0C0C0}8} & -0.18 & \cellcolor[HTML]{C0C0C0}302 & 0.03 & \cellcolor[HTML]{C0C0C0}79 & -2.01 & \cellcolor[HTML]{C0C0C0}409 & -0.12 & \cellcolor[HTML]{C0C0C0}23 \\
\textbf{Snow} & -0.14 & \cellcolor[HTML]{C0C0C0}698 & -0.03 & \cellcolor[HTML]{C0C0C0}67 & -0.02 & \cellcolor[HTML]{C0C0C0}66 & -0.02 & \multicolumn{1}{c|}{\cellcolor[HTML]{C0C0C0}5} & -0.56 & \cellcolor[HTML]{C0C0C0}255 & 0.04 & \cellcolor[HTML]{C0C0C0}68 & -0.84 & \cellcolor[HTML]{C0C0C0}291 & -0.08 & \cellcolor[HTML]{C0C0C0}23 \\
\textbf{Spatter} & -1.06 & \cellcolor[HTML]{C0C0C0}604 & -0.02 & \cellcolor[HTML]{C0C0C0}43 & 0.08 & \cellcolor[HTML]{C0C0C0}52 & 0.02 & \multicolumn{1}{c|}{\cellcolor[HTML]{C0C0C0}4} & -0.29 & \cellcolor[HTML]{C0C0C0}228 & 0.06 & \cellcolor[HTML]{C0C0C0}64 & -0.38 & \cellcolor[HTML]{C0C0C0}239 & -0.01 & \cellcolor[HTML]{C0C0C0}15 \\
\textbf{Speckle\_noise} & -1.89 & \cellcolor[HTML]{C0C0C0}631 & -0.12 & \cellcolor[HTML]{C0C0C0}49 & -0.06 & \cellcolor[HTML]{C0C0C0}65 & 0.01 & \multicolumn{1}{c|}{\cellcolor[HTML]{C0C0C0}5} & -0.34 & \cellcolor[HTML]{C0C0C0}269 & -0.01 & \cellcolor[HTML]{C0C0C0}70 & -1.98 & \cellcolor[HTML]{C0C0C0}398 & -0.04 & \cellcolor[HTML]{C0C0C0}13 \\
\textbf{Zoom\_blur} & 0.6 & \cellcolor[HTML]{C0C0C0}883 & -0.07 & \cellcolor[HTML]{C0C0C0}56 & 0.1 & \cellcolor[HTML]{C0C0C0}77 & -0.04 & \multicolumn{1}{c|}{\cellcolor[HTML]{C0C0C0}12} & 0.25 & \cellcolor[HTML]{C0C0C0}415 & -0.04 & \cellcolor[HTML]{C0C0C0}125 & 1.98 & \cellcolor[HTML]{C0C0C0}392 & 0.01 & \cellcolor[HTML]{C0C0C0}24 \\ \hline
\textbf{|Average|} & 0.31 & \cellcolor[HTML]{C0C0C0}232 & 0.08 & \cellcolor[HTML]{C0C0C0}53 & 0.07 & \cellcolor[HTML]{C0C0C0}58 & 0.03 & \multicolumn{1}{c|}{\cellcolor[HTML]{C0C0C0}9} & 0.20 & \cellcolor[HTML]{C0C0C0}295 & 0.06 & \cellcolor[HTML]{C0C0C0}75 & 0.89 & \cellcolor[HTML]{C0C0C0}307 & 0.05 & \cellcolor[HTML]{C0C0C0}19 \\ \hline
\end{tabular}
}
\end{table}


\begin{table}
\caption{Behavior of quantized models under natural distribution shift. Non-highlighted value: accuracy change (\%), \hl{highlighted value}: number of disagreements. A low value indicates a small difference between the original and quantized models. ID refers to the ID test data and the others are OOD test data. TF: TensorFlowLite. CM: CoreML. |Average|: the average of absolute changes.}
\label{tab:rq1_nds}
\centering
\resizebox{\columnwidth}{!}{
\begin{tabular}{lcccccccccccccccc}
\hline
 & \multicolumn{16}{c}{\textbf{iWildCam}} \\ \cline{2-17} 
 & \multicolumn{8}{c}{\textbf{DenseNet-121}} & \multicolumn{8}{c}{\textbf{ResNet50}} \\
\multirow{-3}{*}{\textbf{Test Data}} & \multicolumn{2}{c}{\textbf{TF-8}} & \multicolumn{2}{c}{\textbf{TF-16}} & \multicolumn{2}{c}{\textbf{CM-8}} & \multicolumn{2}{c}{\textbf{CM-16}} & \multicolumn{2}{c}{\textbf{TF-8}} & \multicolumn{2}{c}{\textbf{TF-16}} & \multicolumn{2}{c}{\textbf{CM-8}} & \multicolumn{2}{c}{\textbf{CM-16}} \\ \hline
\textbf{ID} & -18.96 & \cellcolor[HTML]{C0C0C0}2830 & -0.12 & \cellcolor[HTML]{C0C0C0}167 & -8.34 & \cellcolor[HTML]{C0C0C0}2035 & 0.04 & \multicolumn{1}{c|}{\cellcolor[HTML]{C0C0C0}34} & 0 & \cellcolor[HTML]{C0C0C0}326 & -0.11 & \cellcolor[HTML]{C0C0C0}187 & -0.21 & \cellcolor[HTML]{C0C0C0}226 & 0.06 & \cellcolor[HTML]{C0C0C0}16 \\
\textbf{OOD} & -10.91 & \cellcolor[HTML]{C0C0C0}14279 & -0.42 & \cellcolor[HTML]{C0C0C0}1105 & -5.18 & \cellcolor[HTML]{C0C0C0}11095 & 0 & \multicolumn{1}{c|}{\cellcolor[HTML]{C0C0C0}216} & 1.09 & \cellcolor[HTML]{C0C0C0}2158 & 0.6 & \cellcolor[HTML]{C0C0C0}1270 & -0.33 & \cellcolor[HTML]{C0C0C0}1811 & -0.01 & \cellcolor[HTML]{C0C0C0}128 \\ \hline
\textbf{|Average|} & 14.94 & \cellcolor[HTML]{C0C0C0}8555 & 0.27 & \cellcolor[HTML]{C0C0C0}636 & 6.76 & \cellcolor[HTML]{C0C0C0}6565 & 0.02 & \multicolumn{1}{c|}{\cellcolor[HTML]{C0C0C0}125} & 0.55 & \cellcolor[HTML]{C0C0C0}1242 & 0.35 & \cellcolor[HTML]{C0C0C0}729 & 0.27 & \cellcolor[HTML]{C0C0C0}1019 & 0.04 & \cellcolor[HTML]{C0C0C0}72 \\ \hline
\multicolumn{17}{c}{\textbf{IMDb}} \\ \cline{2-17} 
 & \multicolumn{8}{c}{\textbf{LSTM}} & \multicolumn{8}{c}{\textbf{GRU}} \\
 & \multicolumn{2}{c}{\textbf{TF-8}} & \multicolumn{2}{c}{\textbf{TF-16}} & \multicolumn{2}{c}{\textbf{CM-8}} & \multicolumn{2}{c}{\textbf{CM-16}} & \multicolumn{2}{c}{\textbf{TF-8}} & \multicolumn{2}{c}{\textbf{TF-16}} & \multicolumn{2}{c}{\textbf{CM-8}} & \multicolumn{2}{c}{\textbf{CM-16}} \\ \hline
\textbf{ID} & - & \cellcolor[HTML]{C0C0C0}- & -0.08 & \cellcolor[HTML]{C0C0C0}8 & 0 & \cellcolor[HTML]{C0C0C0}6 & 0 & \multicolumn{1}{c|}{\cellcolor[HTML]{C0C0C0}0} & - & \cellcolor[HTML]{C0C0C0}- & -0.06 & \cellcolor[HTML]{C0C0C0}3 & 0.04 & \cellcolor[HTML]{C0C0C0}2 & 0 & \cellcolor[HTML]{C0C0C0}0 \\
\textbf{CR} & - & \cellcolor[HTML]{C0C0C0}- & -0.04 & \cellcolor[HTML]{C0C0C0}8 & -0.06 & \cellcolor[HTML]{C0C0C0}9 & 0 & \multicolumn{1}{c|}{\cellcolor[HTML]{C0C0C0}0} & - & \cellcolor[HTML]{C0C0C0}- & 0.28 & \cellcolor[HTML]{C0C0C0}30 & 0.2 & \cellcolor[HTML]{C0C0C0}24 & -0.02 & \cellcolor[HTML]{C0C0C0}1 \\
\textbf{Yelp} & - & \cellcolor[HTML]{C0C0C0}- & -0.12 & \cellcolor[HTML]{C0C0C0}14 & 0.02 & \cellcolor[HTML]{C0C0C0}7 & 0 & \multicolumn{1}{c|}{\cellcolor[HTML]{C0C0C0}0} & - & \cellcolor[HTML]{C0C0C0}- & 0.06 & \cellcolor[HTML]{C0C0C0}9 & 0.08 & \cellcolor[HTML]{C0C0C0}8 & 0 & \cellcolor[HTML]{C0C0C0}0 \\ \hline
\textbf{|Average|} & - & \cellcolor[HTML]{C0C0C0}- & 0.08 & \cellcolor[HTML]{C0C0C0}10 & 0.03 & \cellcolor[HTML]{C0C0C0}7 & 0.00 & \multicolumn{1}{c|}{\cellcolor[HTML]{C0C0C0}0} & - & \cellcolor[HTML]{C0C0C0}- & 0.13 & \cellcolor[HTML]{C0C0C0}14 & 0.11 & \cellcolor[HTML]{C0C0C0}11 & 0.01 & \cellcolor[HTML]{C0C0C0}0 \\ \hline
\end{tabular}
}
\end{table}

Concerning the disagreement, even if the quantized model maintains accuracy, there may exist disagreements. For example, in the case of \emph{LeNet1, CM-8}, the accuracy change is 0, but the number of disagreements is 6. Even worse, in \emph{DenseNet-121, CM-16}, 216 disagreements appear without any accuracy change. This calls for the attention that the behaviors of quantized models can not be exactly reflected by only comparing the test accuracy. Thus, during deployment, using accuracy only to evaluate the quality and reliability of quantized DNNs is insufficient.\finding{\textbf{Finding 2}: Disagreements may exist even if quantized models maintain the accuracy.}


Moreover, comparing the number of disagreements from the ID test data and OOD test data, we observe that the distribution shift tends to lead to more disagreements. In 82\% cases (241 of 294), the number of disagreements from OOD test data is greater than from ID test data, the difference can be by up to 5.28\% (LeNet1, Fog, TF-8). However, after the model has been deployed and used in the wild, test data are more likely to have distribution shifts which raises a big concern that model quantization may bring unexcepted errors. \finding{\textbf{Finding 3}: Model quantization is sensitive to the distribution shift where more disagreements happen.}


Next, we compare the two quantization techniques concerning the accuracy change. On average, regardless of the dataset, DNN, and quantization level, CoreML produces more stable quantized models (smaller change) than TensorFlowLite in most cases (12 out of 14). Concretely, in 16-bit float quantization, CoreML always outperforms TensorFlowLite. Take \emph{iWildCam, DenseNet-121} as an example, in 16-bit level quantization, the average accuracy change is 0.27\% by TensorFlowLite but only 0.02\% by CoreML. This difference 0.25\% could cause the CoreML-quantized model to correctly predict 188 more data than TensorFlowLite-quantized model, which is a considerable difference. In 8-bit integer quantization, CoreML can still outperform TensorFlowLite  in most cases (4 out of 6). Additionally, we found an extreme case (\emph{iWildCam, DensetNet-121}) where the accuracy of quantized models by both techniques drops a lot. This finding raises the concern that both quantization tools have room for improvement and require a thorough test. On the other hand, concerning the number of disagreements, the models quantized by CoreML have fewer disagreement inputs than those by TensorFlowLite in most cases (13 out of 14). \finding{\textbf{Finding 4}: In post-training model quantization, compared to TensorFlowLite, CoreML maintains the accuracy better as well as causes fewer disagreements.}


\noindent\colorbox{gray!20}{\framebox{\parbox{0.96\linewidth}{
\textbf{Answer to RQ1}: Under synthetic distribution shift, the accuracy change and number of disagreements between the original and quantized models increase  by up to 3.03\% and 5.28\%. Regardless of the dataset, DNN, and distribution shift, CoreML keeps the behaviors of original DNNs better than TensorFlowLite during deployment.}}}

\subsection{RQ2: Influence of Training Strategy}
\label{subsec:rq2}
In this section, we explore how different training strategies influence the behaviors of quantized models. Due to the space limitation, we only report the results of one model from each dataset (MNIST-LeNet5, CIFAR-10-ResNet20, IMDb-LSTM, and iWildsCam-ResNet50). The whole results are available at our project site.

\begin{figure}[!ht]
    \centering
    \subfigure[MNIST]{
    \includegraphics[scale=0.18]{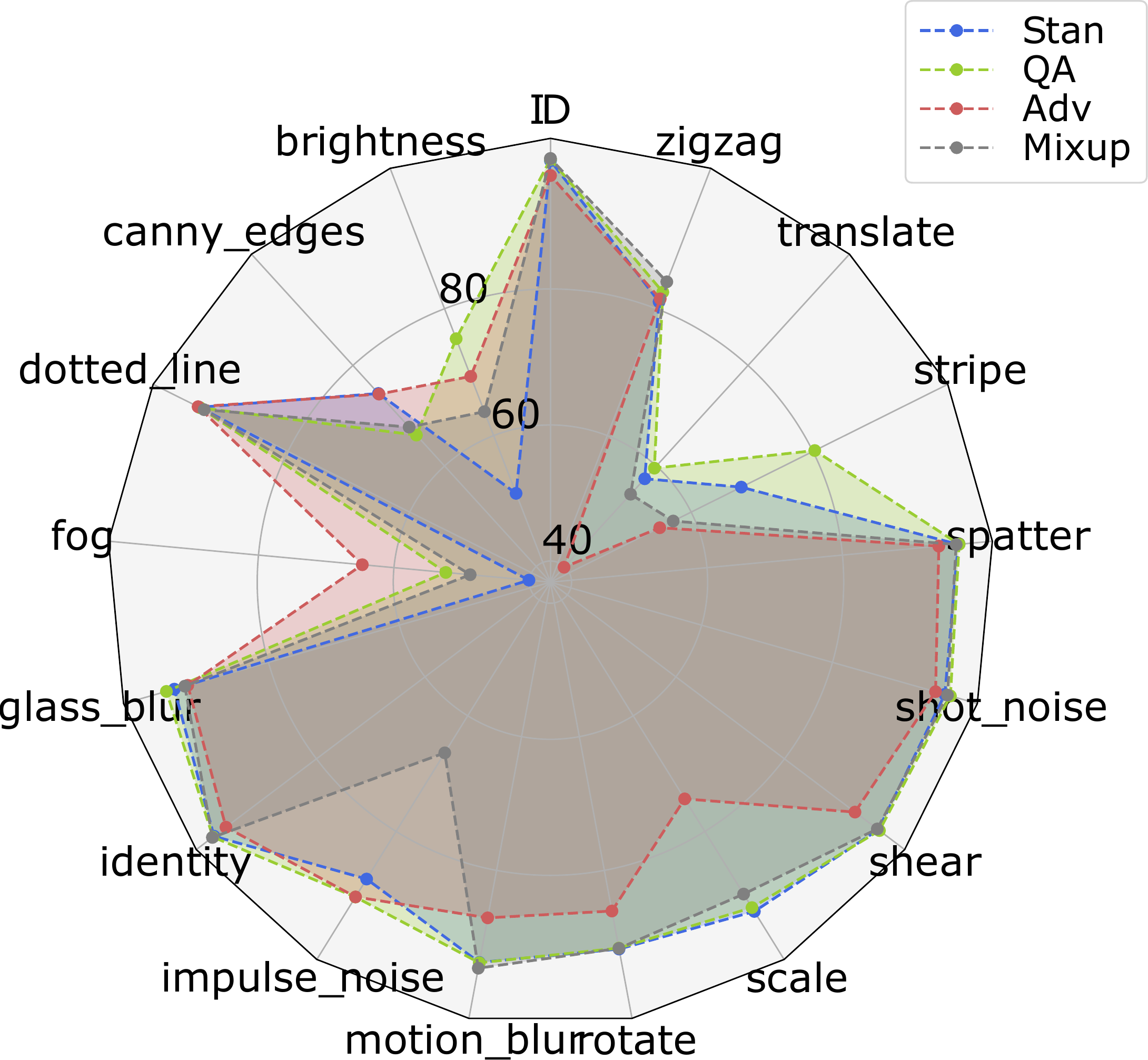}%
    }
    \subfigure[CIFAR-10]{
    \includegraphics[scale=0.18]{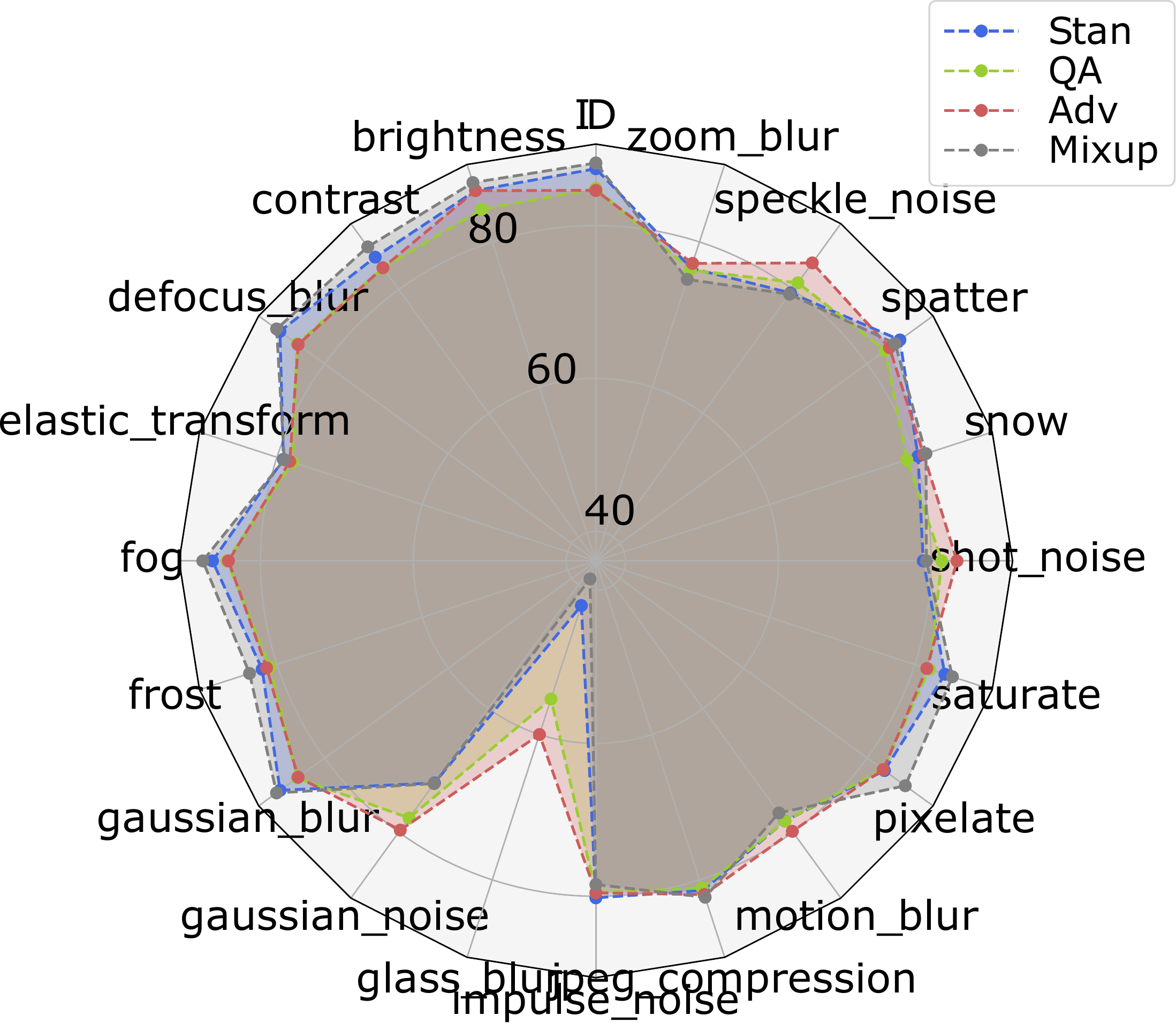}%
    }
    \subfigure[IMDb]{
    \includegraphics[scale=0.185]{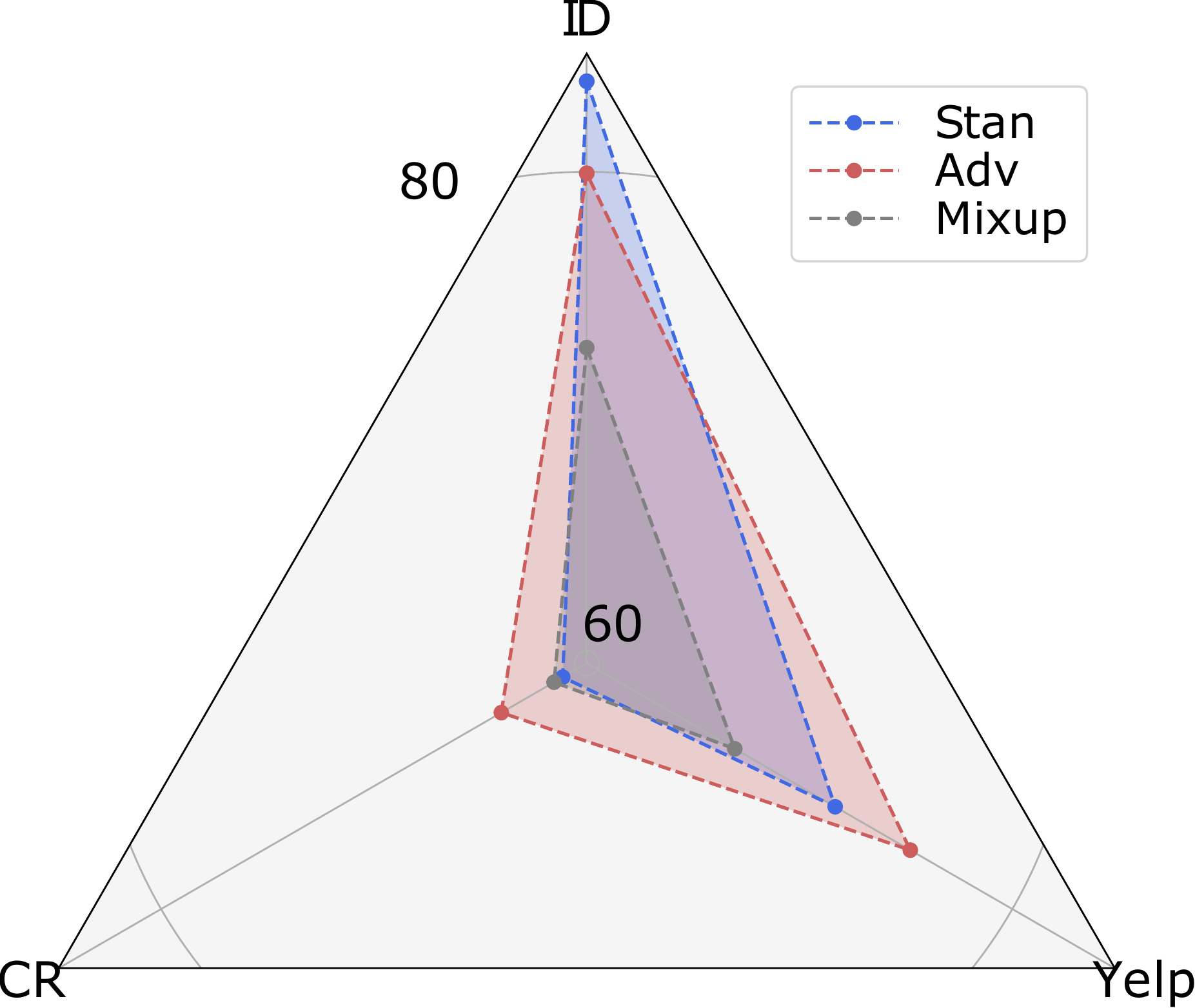}%
    }
    \subfigure[iWildCam]{
    \includegraphics[scale=0.22]{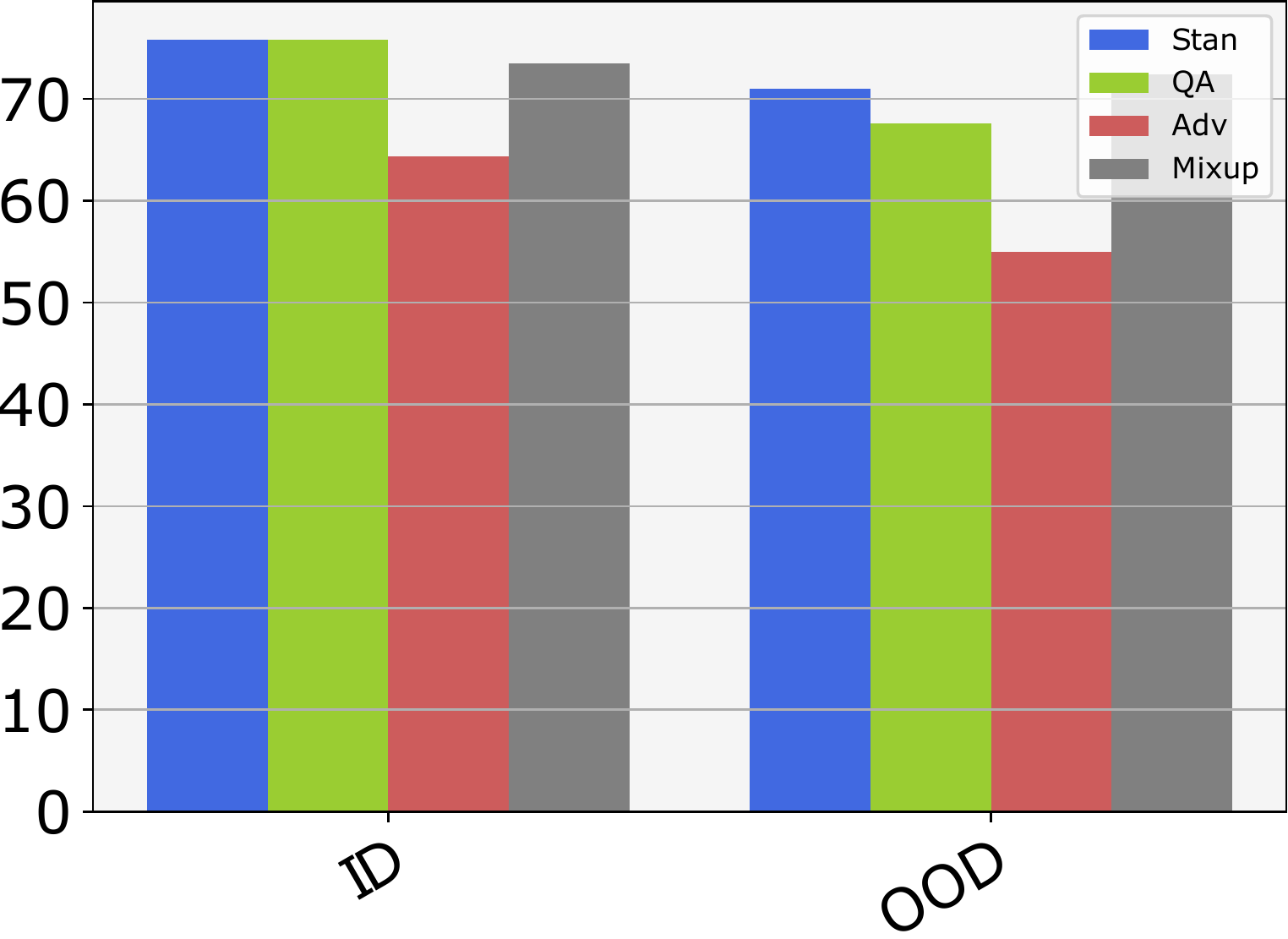}%
    }
    \caption{Accuracy (\%) of models (before quantization) trained by different training strategies. ID represents the accuracy on ID test datasets, and the others are on OOD test datasets. Stan: standard training. QA: quantization-aware training. Adv: adversarial training. Mixup: Mixup training.}
    \label{fig:rq2_acc}
\end{figure}

\begin{figure*}[h]
    \centering
    \subfigure[MNIST, TensorFlowLite]{
    \includegraphics[scale=0.25]{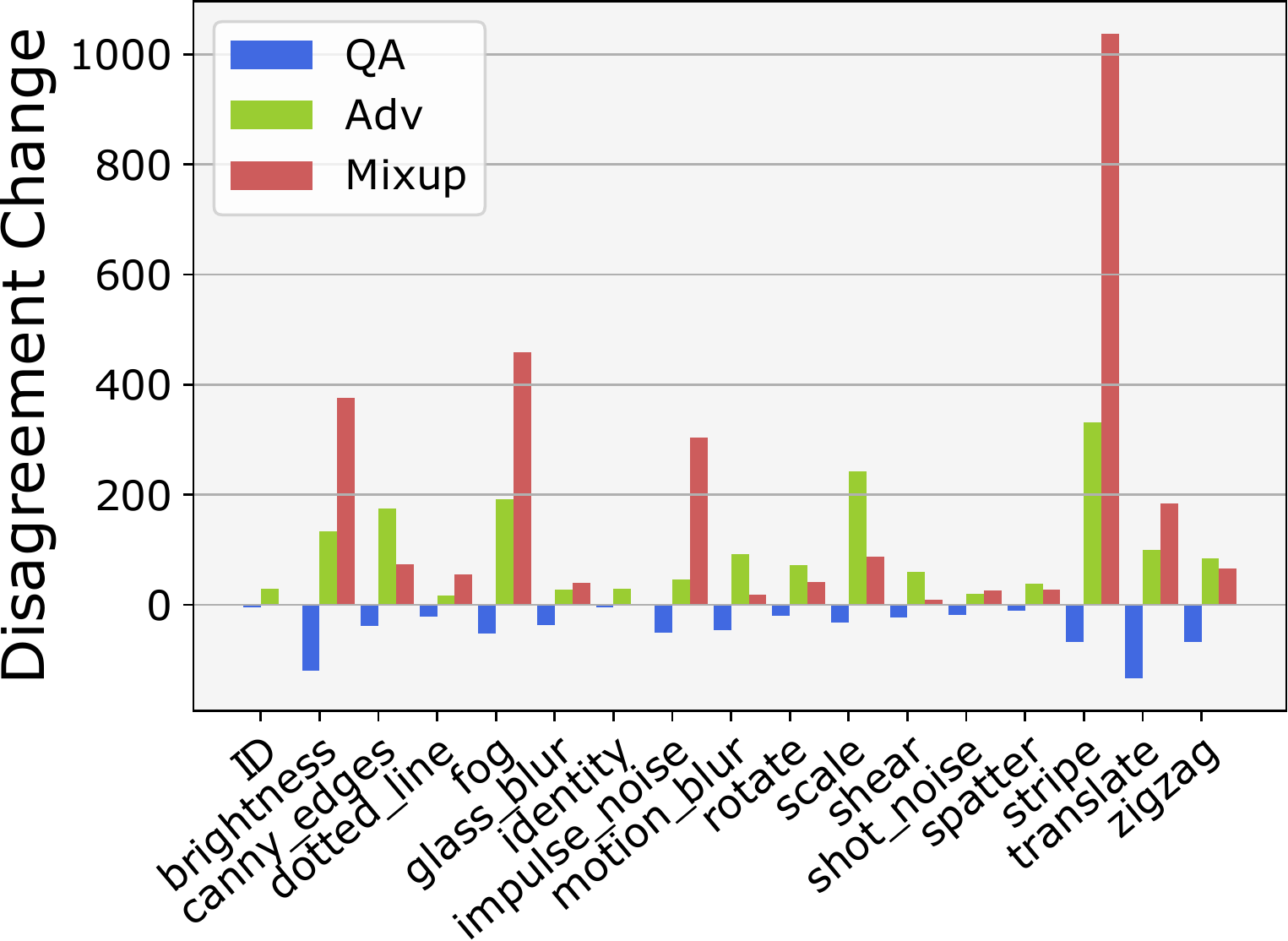}%
    }
    \subfigure[MNIST, CoreML]{
    \includegraphics[scale=0.25]{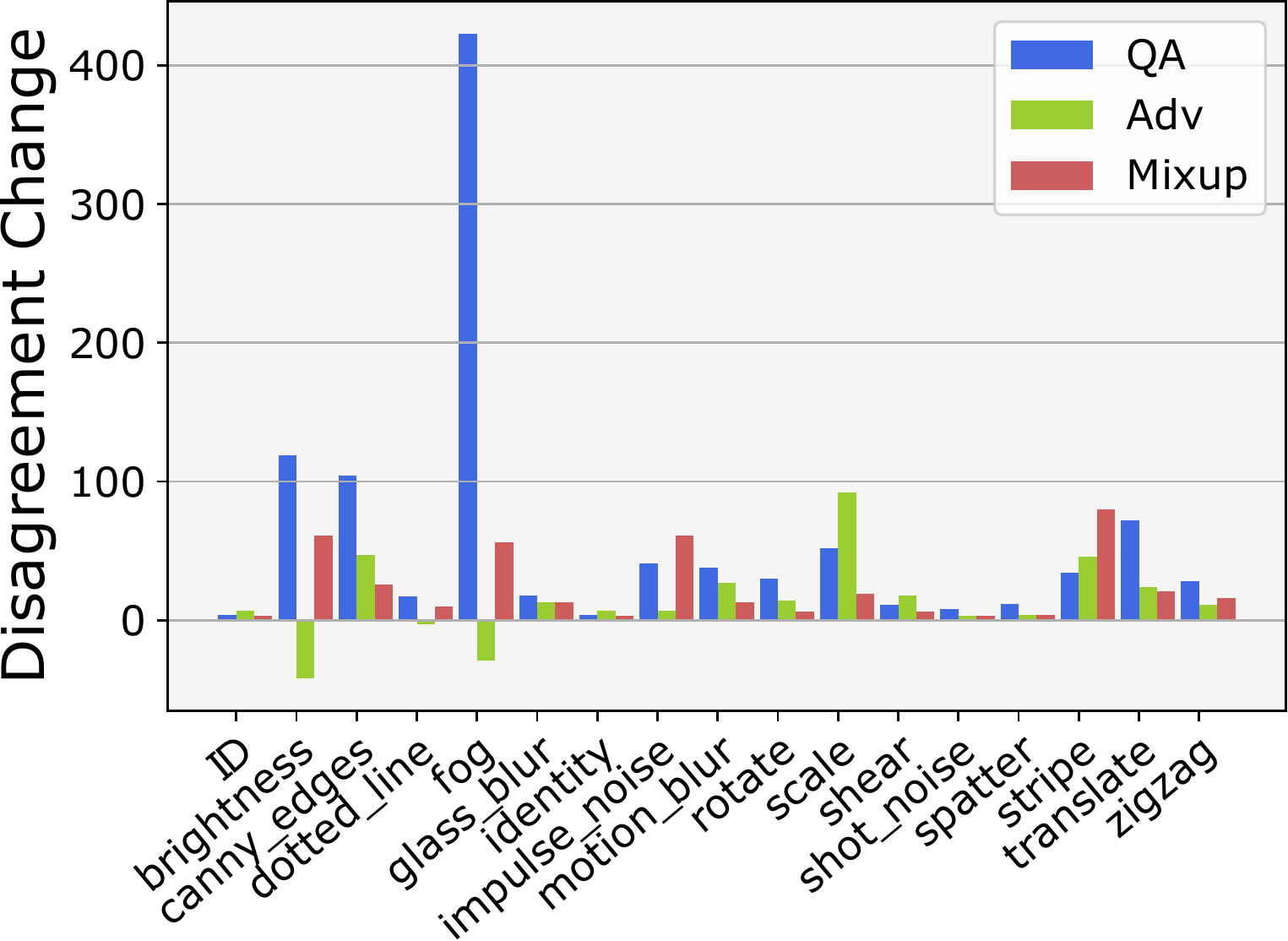}%
    }
    \subfigure[CIFAR-10, TensorFlowLite]{
    \includegraphics[scale=0.25]{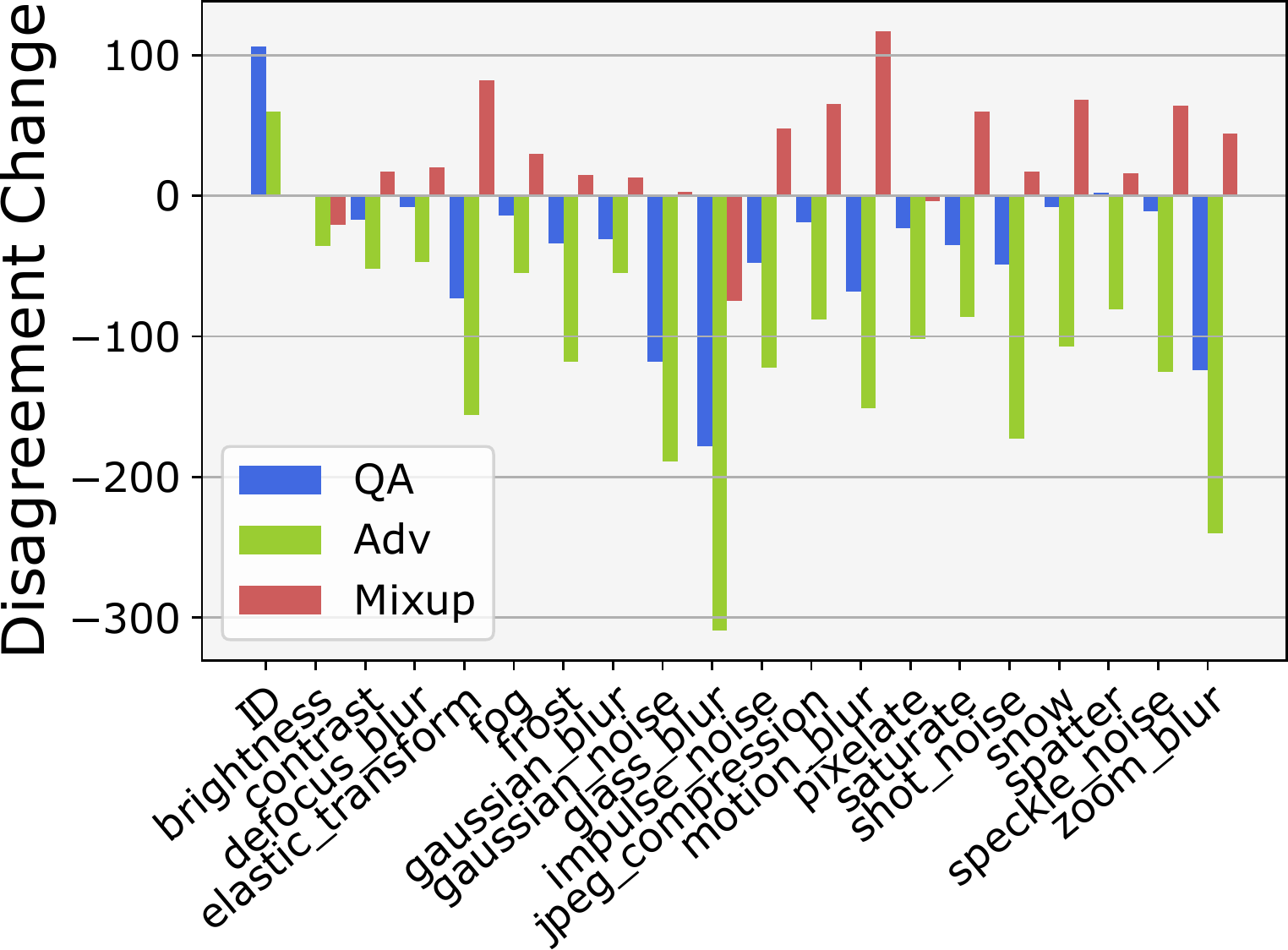}%
    }
    \subfigure[CIFAR-10, CoreML]{
    \includegraphics[scale=0.25]{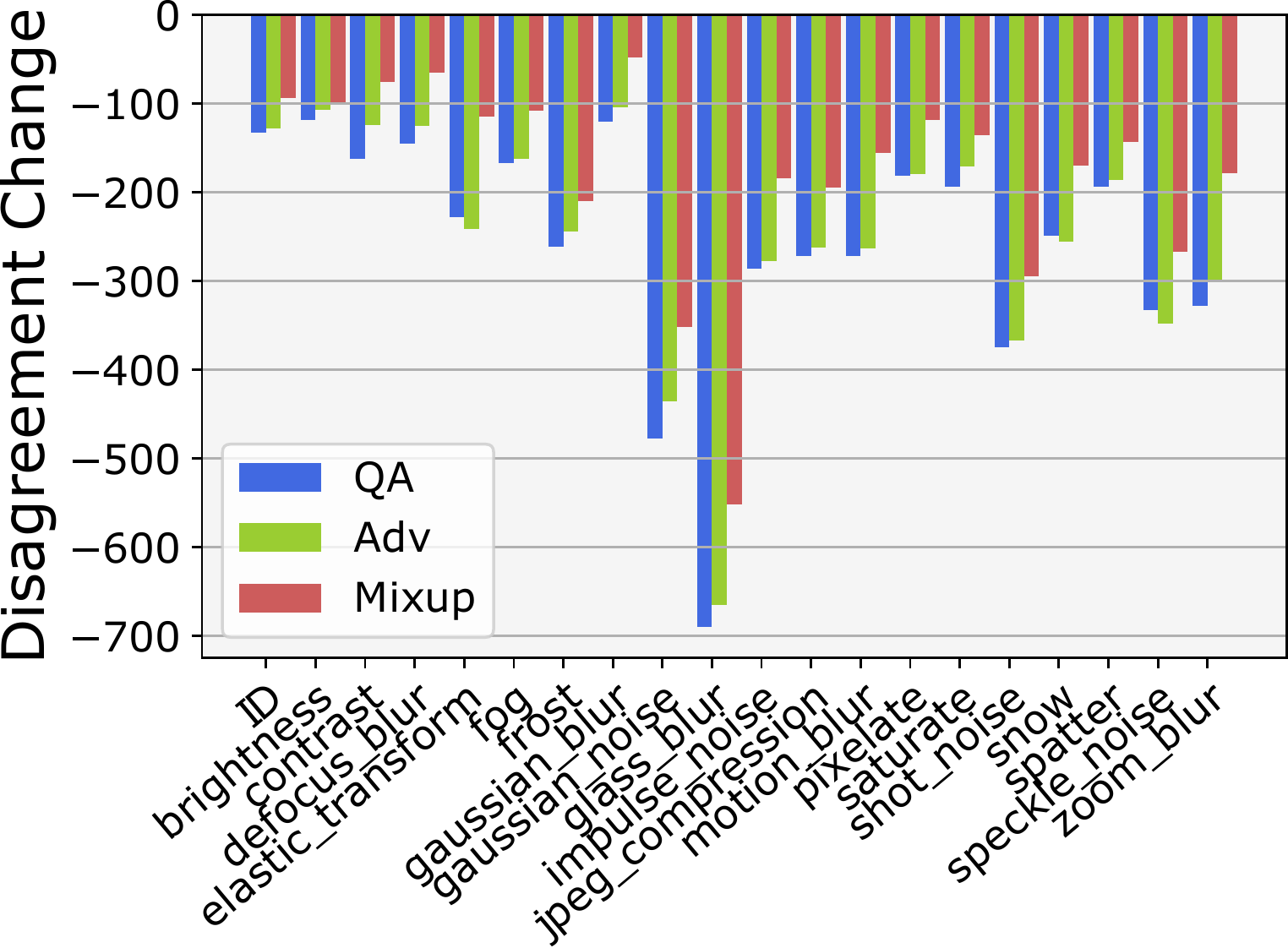}%
    }
    \subfigure[IMDb, TensorFlowLite]{
    \label{fig:dim}
    \includegraphics[scale=0.25]{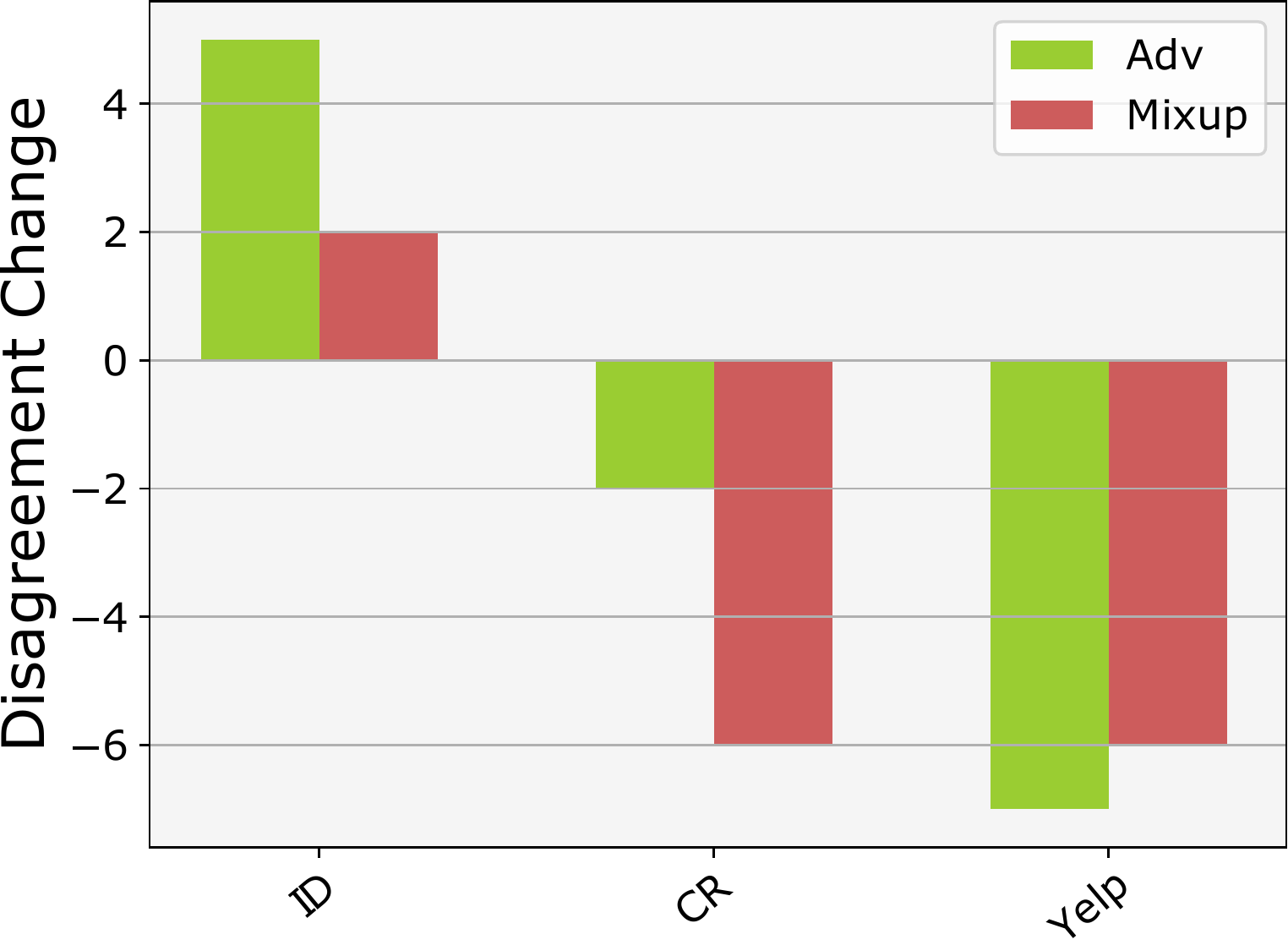}%
    }
    \subfigure[IMDb, CoreML]{
    \includegraphics[scale=0.25]{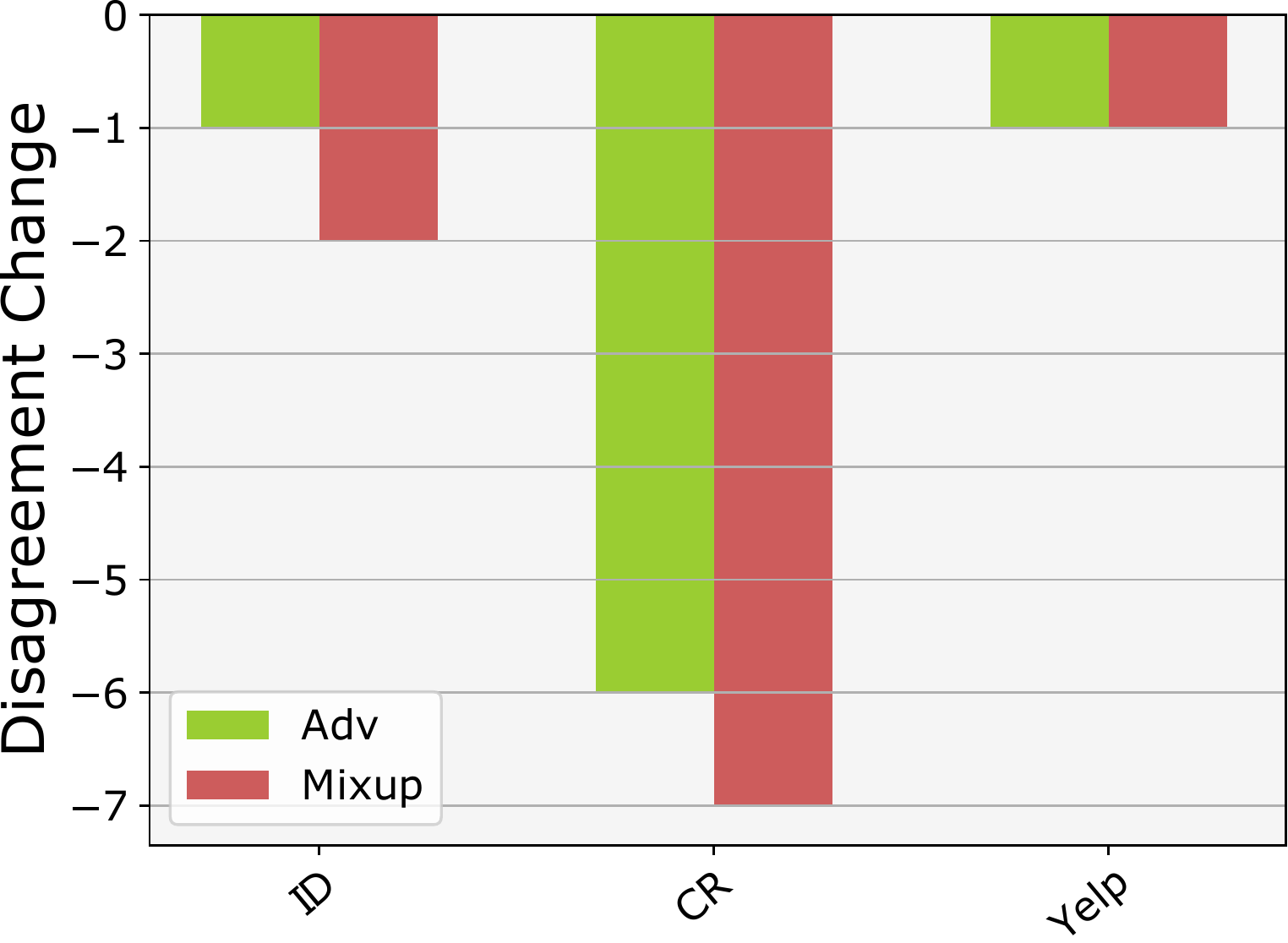}%
    }
    \subfigure[iWildCam, TensorFlowLite]{
    \includegraphics[scale=0.25]{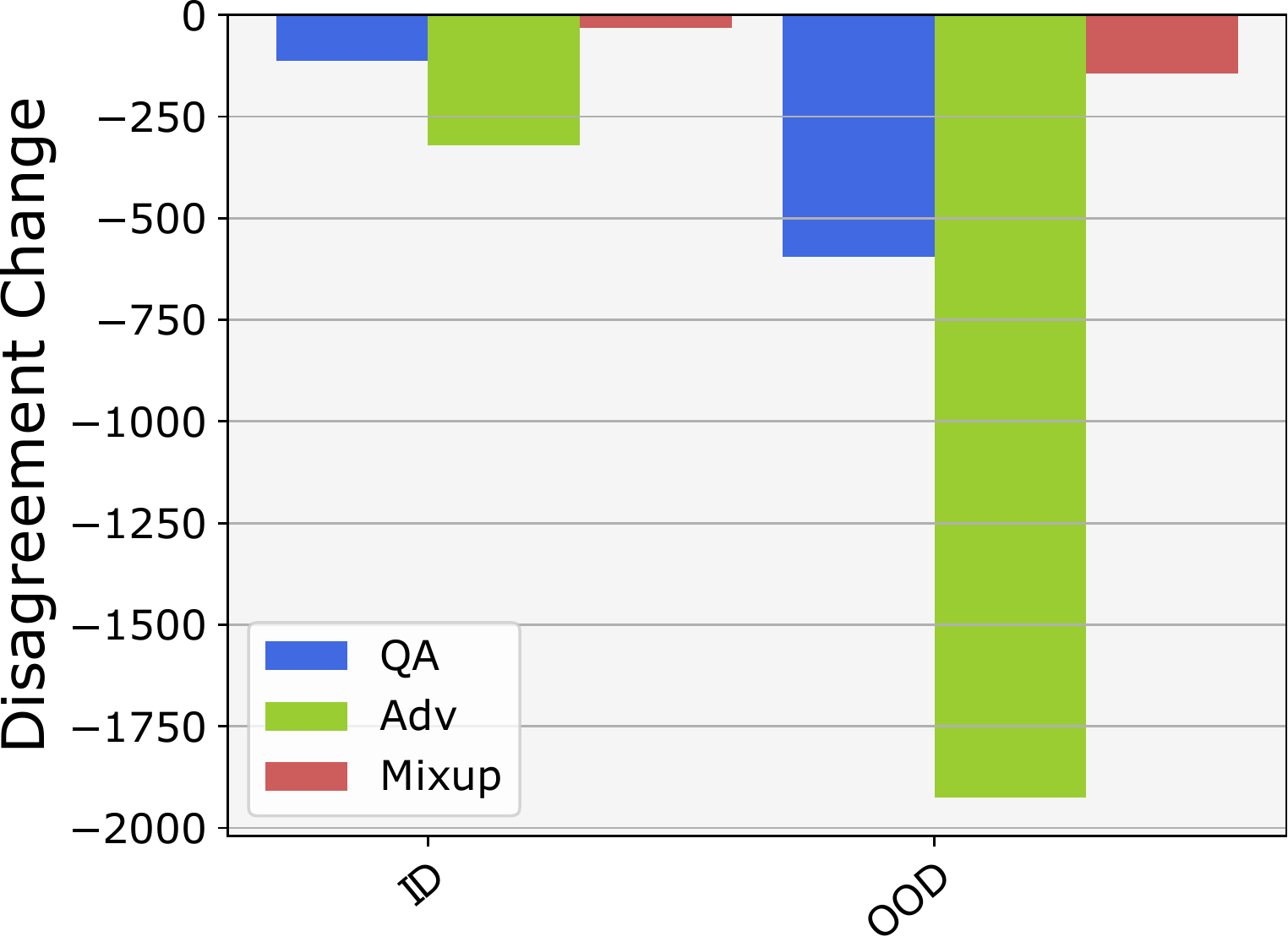}%
    }
    \subfigure[iWildCam, CoreML]{
    \label{fig:diw}
    \includegraphics[scale=0.25]{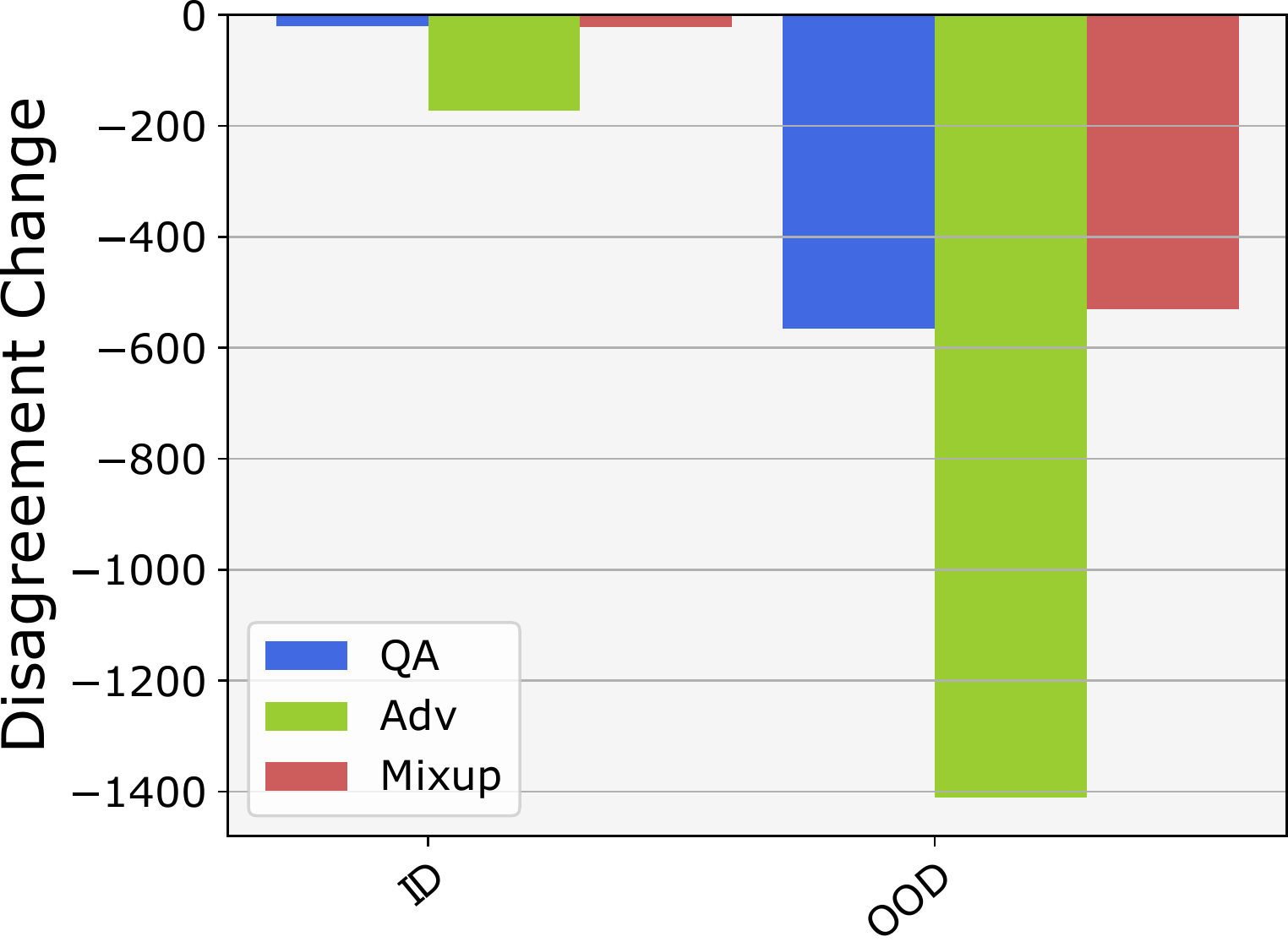}%
    }
    \caption{The disagreement change of models trained by different training strategies compared to by standard training. ID: ID test datasets, and the others are OOD test datasets. QA: quantization-aware training. Adv: adversarial training. Mixup: Mixup training. $y-$axis: the difference of the number of disagreements between a training strategy and the standard training.}
    \label{fig:rq2_dis}
\end{figure*}

First, we evaluate the performance of each training strategy concerning the distribution shift before model quantization. Figure \ref{fig:rq2_acc} shows the results. Under synthetic distribution shift, for MNIST, there are 12, 5, 6 cases out of 17 that using quantization-aware, adversarial, and Mixup training, respectively, improve the accuracy compared to using standard training. While the result for CIFAR-10 changes to 5, 8, and 12 cases of 20 correspondingly. We conclude that none of these three training strategies can consistently deal with the issue of accuracy degradation under synthetic distribution shifts. On the other hand, under natural distribution shift, interestingly, when performing adversarial training for IMDb models, the accuracy of models on both distribution shifted datasets ($CR$ and $Yelp$) has been improved. We conjecture that the features of text adversarial examples are more likely to appear in the OOD test dataset. For example, the original sentence \textit{"a wonderful...are terribly \textbf{well done}} and its adversarial sentence \textit{"a wonderful...are terribly \textbf{considerably perform}"} only have two-word difference, but the model predicts differently. We found that the words \textit{considerably} and \textit{perform} are both in the vocabulary of OOD data. For iWildCam, only the Mixup training can improve the accuracy of models on shifted data. \finding{\textbf{Finding 5}: Concerning the accuracy, none of the three (quantization-aware, adversarial, Mixup) training strategies has a significant advantage over standard training before quantization.}


\begin{table}[]
\caption{Average accuracy change (\%) of models by quantization. The average is of 17, 20, 3, and 2 test datasets of MNIST, CIFAR-10, IMDb, and iWildCam, respectively. A low average value indicates a small difference between the original and quantized models. \hl{Highlighted values} indicate that the accuracy change by the corresponding training strategy is the same as or smaller than by standard training. TF: TensorFlowLite. CM: CoreML. Baseline: standard training.}
\label{tab:rq2_acc_change}
\centering
\setlength{\abovecaptionskip}{3pt}
\scriptsize
\setlength\tabcolsep{2pt}
\resizebox{.85\columnwidth}{!}{
\begin{tabular}{lccccc}
\hline
 & \multicolumn{1}{l}{} & \multicolumn{4}{c}{\textbf{Training Strategy}} \\ \cline{3-6} 
\multirow{-2}{*}{\textbf{Dataset}} & \multicolumn{1}{l}{\multirow{-2}{*}{\textbf{Quantization}}} & \textbf{Standard} & \textbf{Quantization-aware} & \textbf{Adversarial} & \textbf{Mixup} \\ \hline
 & TF-8 & 0.09 & \cellcolor[HTML]{C0C0C0}0.06 & 0.36 & 0.53 \\
 & TF-16 & 0.11 & \cellcolor[HTML]{C0C0C0}0.01 & \cellcolor[HTML]{C0C0C0}0.09 & 0.24 \\
 & CM-8 & 0.11 & \cellcolor[HTML]{C0C0C0}0.06 & \cellcolor[HTML]{C0C0C0}0.11 & \cellcolor[HTML]{C0C0C0}0.09 \\
\multirow{-4}{*}{\textbf{MNIST}} & CM-16 & 0.01 & \cellcolor[HTML]{C0C0C0}0.01 & \cellcolor[HTML]{C0C0C0}0.01 & 0.02 \\ \hline
 & TF-8 & 0.20 & 0.77 & 1.02 & 1.48 \\
 & TF-16 & 0.06 & \cellcolor[HTML]{C0C0C0}0.06 & 0.21 & 0.15 \\
 & CM-8 & 0.89 & \cellcolor[HTML]{C0C0C0}0.05 & \cellcolor[HTML]{C0C0C0}0.20 & \cellcolor[HTML]{C0C0C0}0.12 \\
\multirow{-4}{*}{\textbf{CIFAR-10}} & CM-16 & 0.05 & \cellcolor[HTML]{C0C0C0}0.03 & 0.16 & \cellcolor[HTML]{C0C0C0}0.04 \\ \hline
 & TF-16 & 0.08 & - & \cellcolor[HTML]{C0C0C0}0.03 & \cellcolor[HTML]{C0C0C0}0.05 \\
 & CM-8 & 0.03 & - & \cellcolor[HTML]{C0C0C0}0.03 & \cellcolor[HTML]{C0C0C0}0.03 \\
\multirow{-3}{*}{\textbf{IMDb}} & CM-16 & 0.01 & - & \cellcolor[HTML]{C0C0C0}0.01 & \cellcolor[HTML]{C0C0C0}0.01 \\ \hline
 & TF-8 & 0.55 & \cellcolor[HTML]{C0C0C0}0.47 & \cellcolor[HTML]{C0C0C0}0.24 & \cellcolor[HTML]{C0C0C0}0.29 \\
 & TF-16 & 0.35 & \cellcolor[HTML]{C0C0C0}0.05 & \cellcolor[HTML]{C0C0C0}0.10 & \cellcolor[HTML]{C0C0C0}0.07 \\
 & CM-8 & 0.27 & \cellcolor[HTML]{C0C0C0}0.19 & \cellcolor[HTML]{C0C0C0}0.04 & 0.62 \\
\multirow{-4}{*}{\textbf{iWildCam}} & CM-16 & 0.04 & 0.11 & \cellcolor[HTML]{C0C0C0}0.04 & 0.06 \\ \hline
\end{tabular}
}
\vspace{-4.8mm}
\end{table}

Second, we check the accuracy change of each model trained by different training strategies after quantization. Table \ref{tab:rq2_acc_change} presents the results of the average accuracy change of all test datasets of each model. Compared to standard training, the quantized models by using the quantization-aware training are more stable where the accuracy change in most cases (10 out of 12) is the same as or smaller. For example, in \emph{CIFAR-10, CM-8}, by standard training, the quantized model has an average of 0.89\% difference compared to its original model. However, by quantization-aware training, the difference can decline to only 0.05\%. By contrast, both adversarial and Mixup training can result in more stable (11 out of 15, 8 out of 15 cases) quantized models than standard training but not as well as quantization-aware training. In short, quantization-aware training outperforms adversarial and Mixup training concerning minimizing the accuracy change during deployment. 

In addition, similar to the findings in RQ1, we observe that under synthetic distribution shift (MNIST and CIFAR-10),  most (7 out of 8) of the accuracy change improvements happen in the models quantized by TensorFlowLite. And for the data with natural distribution shifts, the accuracy change increase only happens in the models quantized by CoreML. \finding{\textbf{Finding 6}: In terms of accuracy change, quantization-aware training produces more stable models than standard, adversarial, and Mixup training. During deployment, TensorFlowLite is more suitable to deal with natural distribution shift, while CoreML performs better for synthetic distribution shift.}


Finally, we check the disagreements that occur during model quantization. Figure \ref{fig:rq2_dis} shows disagreement change of models trained by different strategies compared to by the standard training. Given all OOD test datasets, the quantization-aware equipped with TensorFlowLite can efficiently decrease the number of disagreements. Under synthetic distribution shift only, after TensorFlowLite quantization, the models trained by Mixup training happen more disagreements. On the other hand, under natural distribution shift, all these tree training strategies are useful to reduce disagreements (negative disagreement change in Figures \ref{fig:dim} - \ref{fig:diw}) regardless of the quantization technique. \finding{\textbf{Finding 7}: Under synthetic distribution shift, quantization-aware training is useful to remove disagreements for TensorFlowLite-quantized models. While under natural distribution shift, all three training strategies are efficient to reduce disagreements.}


\noindent\colorbox{gray!20}{\framebox{\parbox{0.96\linewidth}{
\textbf{Answer to RQ2}: Generally, quantization-aware training can produce more stable models with small accuracy changes and fewer disagreements after model quantization. For data with natural distribution shifts, both quantization-aware training and basic data augmentation training (adversarial training and Mixup training) can reduce the disagreements.}}}

\subsection{RQ3: Characteristic of Disagreements}
\label{subsec:rq3}
To understand which data are likely to cause disagreements during quantization, we explore the data properties based on the output uncertainty. The intuition is that the disagreements are those close to the decision boundary of the model \cite{xie2019diffchaser}.  Concretely, after quantization, the decision boundary of a model may slightly move due to the precision of parameter change. As a result, the data that are close to the boundary might cross over the boundary and cause disagreements. Generally, those data are uncertain to the model. Many uncertainty metrics have been developed but which one can be used to more precisely distinguish the disagreements and normal inputs is unclear. In our study, we consider four (Entropy, Margin, Gini, Least Confidence) widely used uncertainty metrics only based on the output of the model to determine the best one to present the property of disagreements.


\begin{figure}[!ht]
    \centering
    \subfigure[Entropy]{
    \includegraphics[scale=0.25]{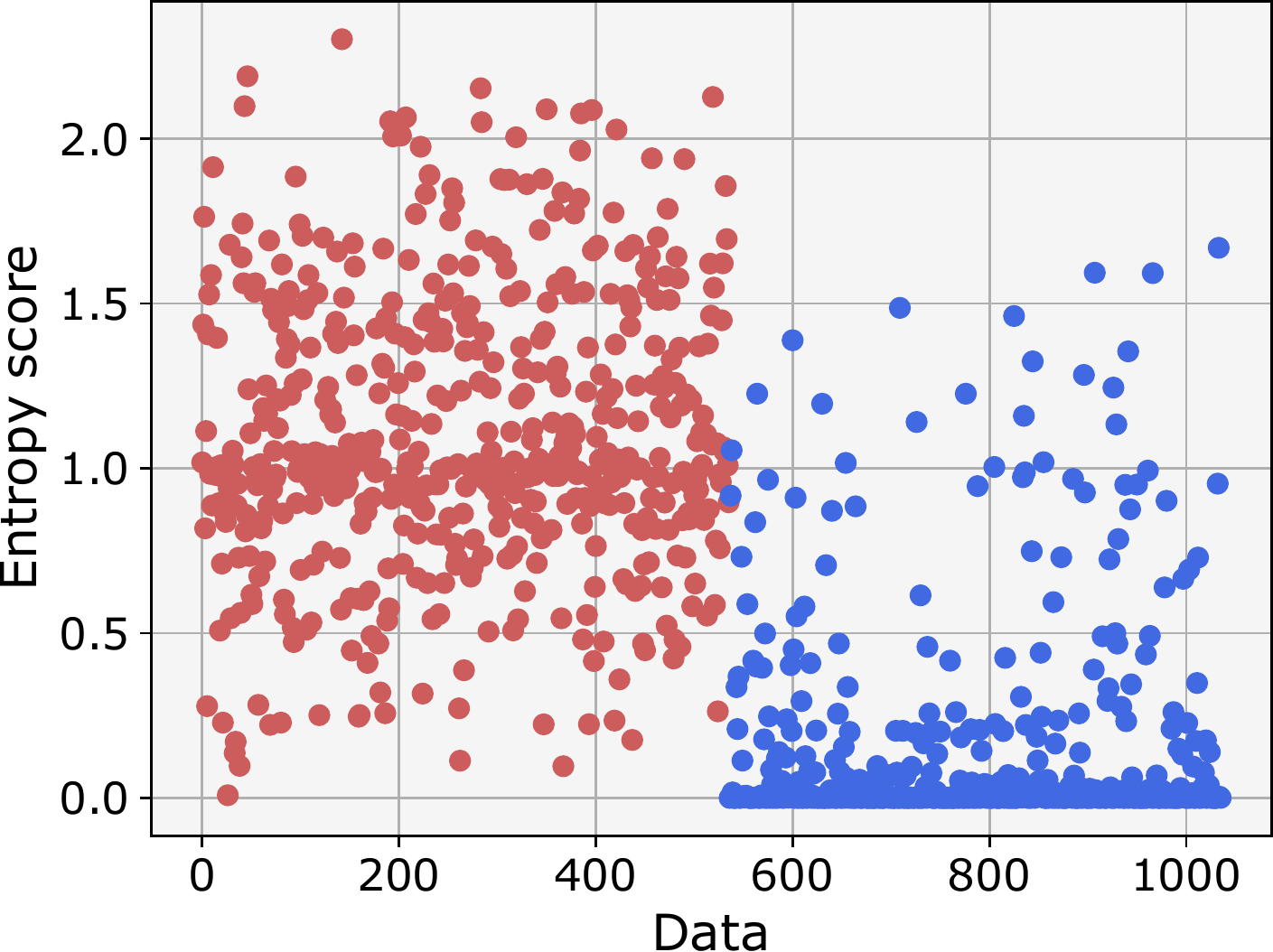}%
    }
    \subfigure[Margin]{
    \includegraphics[scale=0.25]{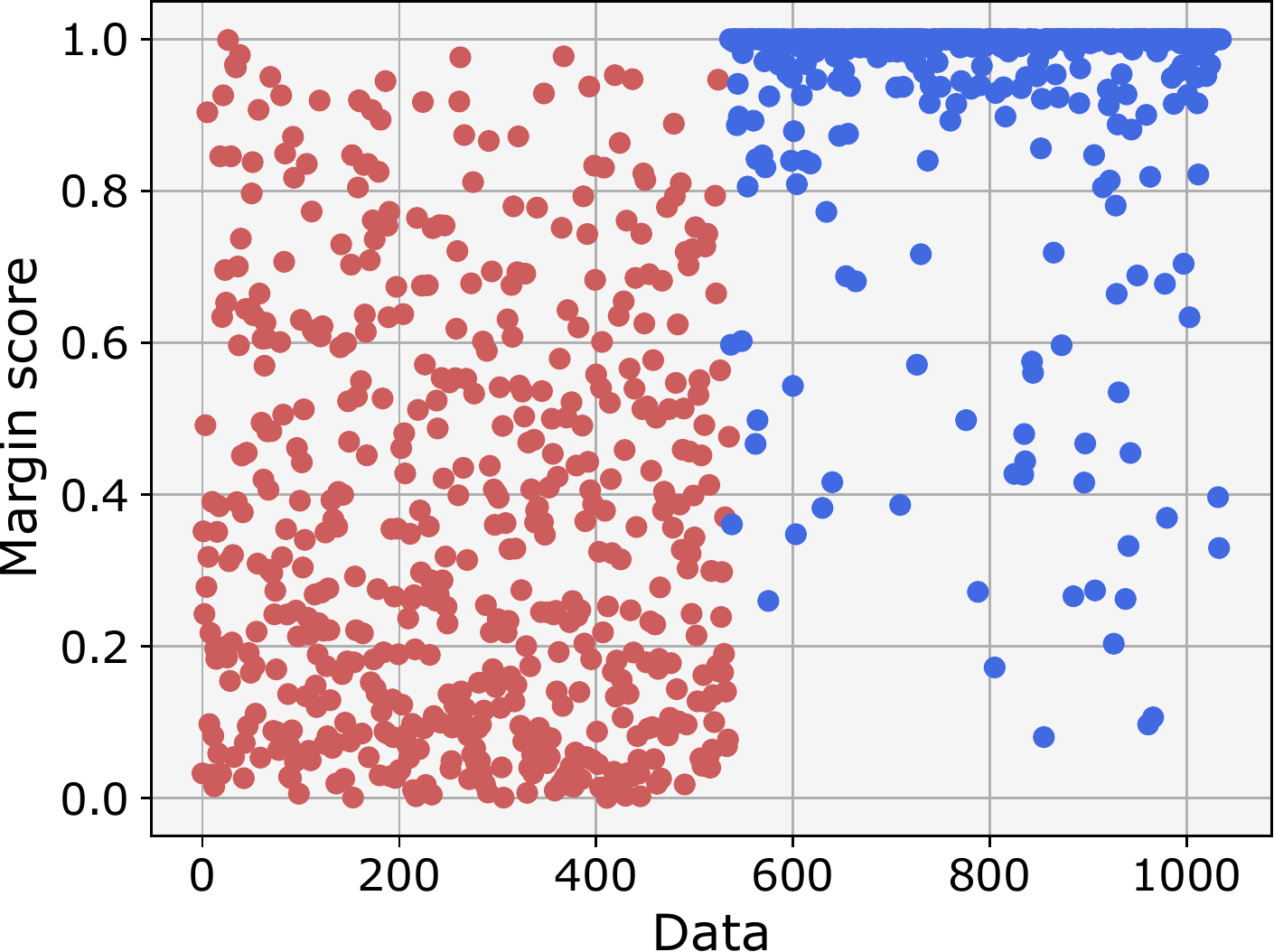}%
    }
    \subfigure[Gini]{
    \includegraphics[scale=0.25]{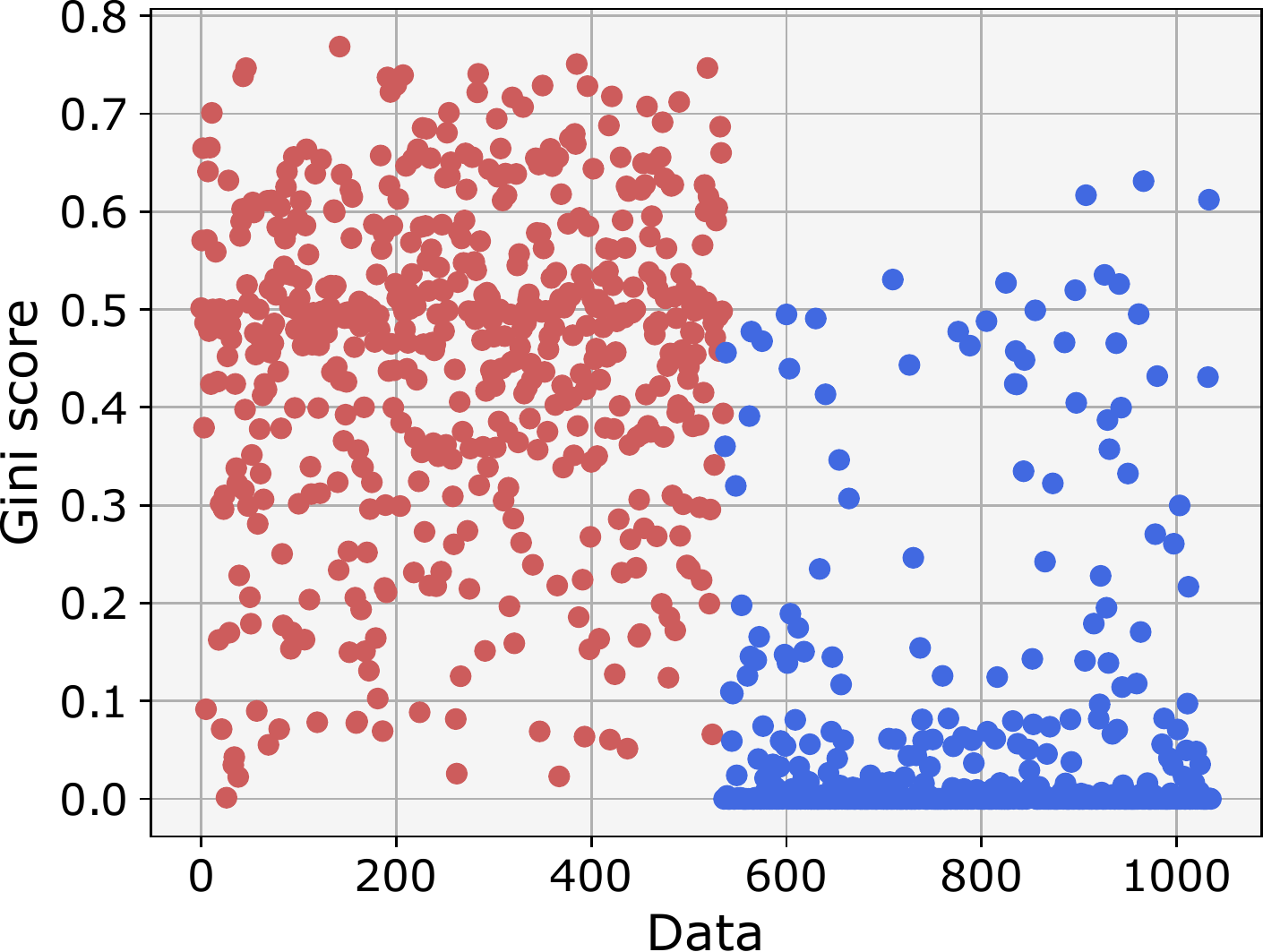}%
    }
    \subfigure[Least Confidence]{
    \includegraphics[scale=0.25]{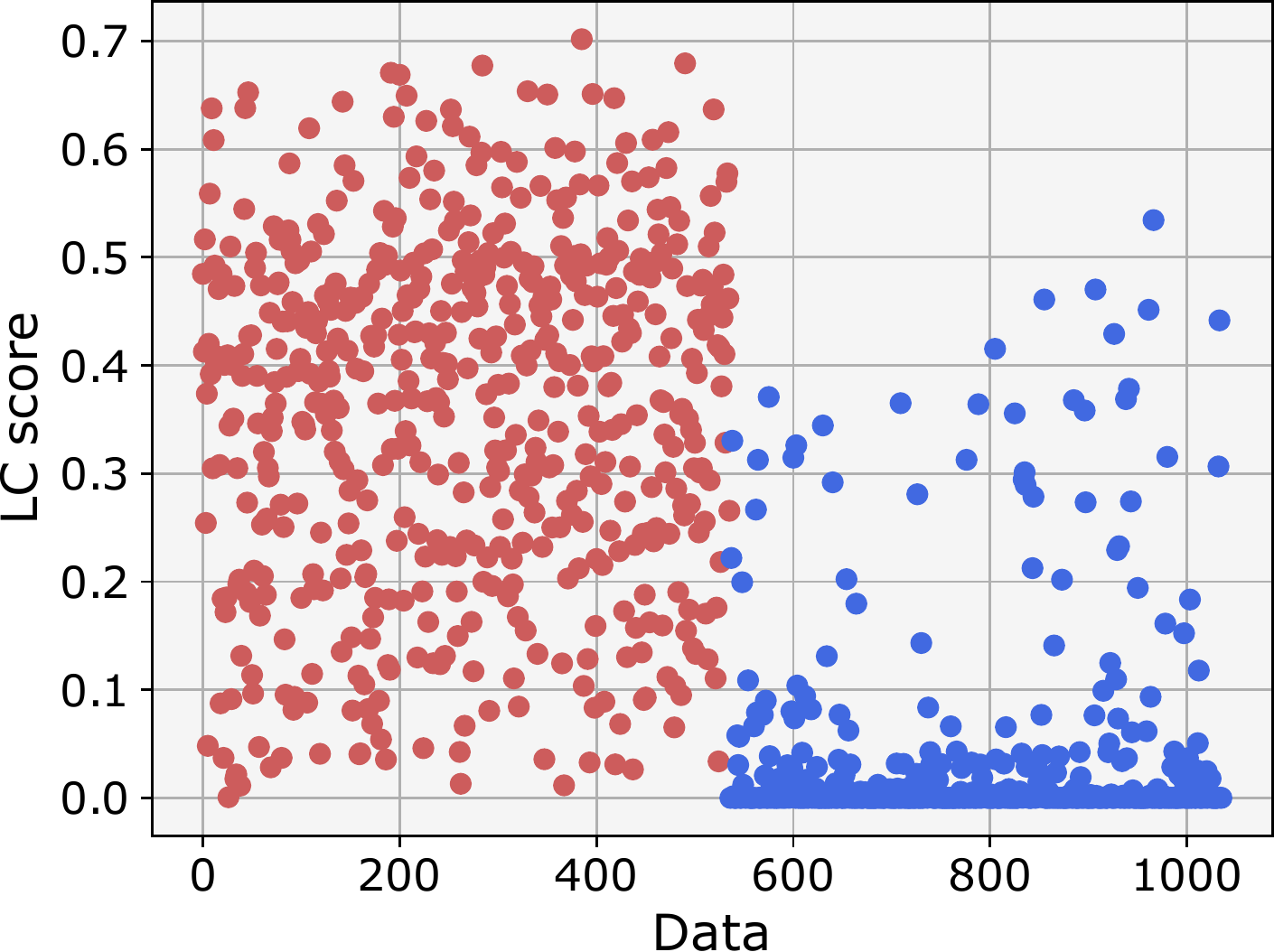}%
    }
    \caption{An example (CIFAR-10, ResNet20, ID test data) of the distributions of output uncertainty scores. Red: disagreements. Blue: normal inputs.}
    \label{fig:rq3_diff}
\end{figure}

Figure \ref{fig:rq3_diff} gives an example (CIFAR-10, ResNet20) of the distribution of uncertainty scores of the disagreements and normal inputs. First of all, regardless of the uncertainty metric, the result confirms that disagreements are more uncertain for a model than normal inputs as they usually have higher (lower in Margin) uncertainty scores. Take the least confidence as an example, most normal inputs have LC scores near 0. According to the definition of LC (in Equation (\ref{eq:lc})), the result demonstrates that the model is confident (with almost 100\%) in the top-1 predictions for these inputs. In detail, the number of inputs having LC scores in the ranges of [0, 0.2], (0.2, 0.4], (0.4, 0.6], (0.6, 0.8], and (0.8, 1] are 462, 31, 7, 0, and 0 respectively. In contrast, for the disagreement inputs, most of them have high uncertain scores. Specifically, the number of inputs that the LC scores in the ranges of [0, 0.2], (0.2, 0.4], (0.4, 0.6], (0.6, 0.8], and (0.8, 1] are 110, 175, 225, 26, and 0 respectively. \finding{\textbf{Finding 8}: Output-based uncertainty is a promising indicator to distinguish disagreements and normal inputs.}


\begin{table}[]
\caption{$AUC-ROC$ score of the logistic regression classifiers trained by using different uncertainty scores. A high value indicates a significant difference between two sets. The best results among four uncertainty metrics are \hl{highlighted}.}
\label{tab:rq3}
\centering
\resizebox{.95\columnwidth}{!}{
\begin{tabular}{ccccccc}
\hline
\multicolumn{1}{l}{} &  & \multicolumn{1}{l}{} & \multicolumn{4}{c}{\textbf{Uncertainty Measure}} \\ \cline{4-7} 
\multicolumn{1}{l}{\multirow{-2}{*}{\textbf{Dataset}}} & \multirow{-2}{*}{\textbf{DNN}} & \multicolumn{1}{l}{\multirow{-2}{*}{\textbf{Training Strategy}}} & \cellcolor[HTML]{FFFFFF}\textbf{Entropy} & \cellcolor[HTML]{FFFFFF}\textbf{Gini} & \cellcolor[HTML]{FFFFFF}\textbf{Margin} & \cellcolor[HTML]{FFFFFF}\textbf{LC} \\ \hline
\rowcolor[HTML]{FFFFFF} 
\cellcolor[HTML]{FFFFFF} & \cellcolor[HTML]{FFFFFF} & Standard & 83.67 & 85.00 & \cellcolor[HTML]{C0C0C0}94.76 & 89.78 \\
\rowcolor[HTML]{FFFFFF} 
\cellcolor[HTML]{FFFFFF} & \cellcolor[HTML]{FFFFFF} & Quantization-aware & 95.81 & 95.20 & \cellcolor[HTML]{C0C0C0}97.45 & 96.61 \\
\rowcolor[HTML]{FFFFFF} 
\cellcolor[HTML]{FFFFFF} & \cellcolor[HTML]{FFFFFF} & Adversarial & 71.41 & 73.58 & \cellcolor[HTML]{C0C0C0}96.51 & 82.49 \\
\rowcolor[HTML]{FFFFFF} 
\cellcolor[HTML]{FFFFFF} & \cellcolor[HTML]{FFFFFF} & Mixup & 79.74 & 84.34 & \cellcolor[HTML]{C0C0C0}94.06 & 89.76 \\
\rowcolor[HTML]{FFFFFF} 
\cellcolor[HTML]{FFFFFF} & \multirow{-5}{*}{\cellcolor[HTML]{FFFFFF}Lenet1} & Average & 82.66 & 84.53 & \cellcolor[HTML]{C0C0C0}95.70 & 89.66 \\ \cline{2-7} 
\rowcolor[HTML]{FFFFFF} 
\cellcolor[HTML]{FFFFFF} & \cellcolor[HTML]{FFFFFF} & \cellcolor[HTML]{FFFFFF}Standard & 86.79 & 89.49 & \cellcolor[HTML]{C0C0C0}97.36 & 94.49 \\
\rowcolor[HTML]{FFFFFF} 
\cellcolor[HTML]{FFFFFF} & \cellcolor[HTML]{FFFFFF} & \cellcolor[HTML]{FFFFFF}Quantization-aware & 80.39 & 83.3 & \cellcolor[HTML]{C0C0C0}94.82 & 88.48 \\
\rowcolor[HTML]{FFFFFF} 
\cellcolor[HTML]{FFFFFF} & \cellcolor[HTML]{FFFFFF} & \cellcolor[HTML]{FFFFFF}Adversarial & 72.02 & 76.47 & \cellcolor[HTML]{C0C0C0}96.78 & 85.09 \\
\rowcolor[HTML]{FFFFFF} 
\cellcolor[HTML]{FFFFFF} & \cellcolor[HTML]{FFFFFF} & \cellcolor[HTML]{FFFFFF}Mixup & 71.53 & 72.00 & \cellcolor[HTML]{C0C0C0}89.42 & 78.37 \\
\rowcolor[HTML]{FFFFFF} 
\multirow{-10}{*}{\cellcolor[HTML]{FFFFFF}\textbf{MNIST}} & \multirow{-5}{*}{\cellcolor[HTML]{FFFFFF}Lenet5} & \cellcolor[HTML]{FFFFFF}Average & 77.68 & 80.32 & \cellcolor[HTML]{C0C0C0}94.60 & 86.61 \\ \hline
\rowcolor[HTML]{FFFFFF} 
\cellcolor[HTML]{FFFFFF} & \cellcolor[HTML]{FFFFFF} & \cellcolor[HTML]{FFFFFF}Standard & \cellcolor[HTML]{C0C0C0}95.42 & 93.58 & 94.54 & 94.28 \\
\rowcolor[HTML]{FFFFFF} 
\cellcolor[HTML]{FFFFFF} & \cellcolor[HTML]{FFFFFF} & \cellcolor[HTML]{FFFFFF}Quantization-aware & 95.29 & 95.62 & \cellcolor[HTML]{C0C0C0}96.52 & 96.11 \\
\rowcolor[HTML]{FFFFFF} 
\cellcolor[HTML]{FFFFFF} & \cellcolor[HTML]{FFFFFF} & \cellcolor[HTML]{FFFFFF}Adversarial & 92.63 & 94.01 & \cellcolor[HTML]{C0C0C0}97.05 & 96.04 \\
\rowcolor[HTML]{FFFFFF} 
\cellcolor[HTML]{FFFFFF} & \cellcolor[HTML]{FFFFFF} & \cellcolor[HTML]{FFFFFF}Mixup & 87.03 & 90.31 & \cellcolor[HTML]{C0C0C0}95.28 & 93.27 \\
\rowcolor[HTML]{FFFFFF} 
\cellcolor[HTML]{FFFFFF} & \multirow{-5}{*}{\cellcolor[HTML]{FFFFFF}ResNet20} & \cellcolor[HTML]{FFFFFF}Average & 92.59 & 93.38 & \cellcolor[HTML]{C0C0C0}95.85 & 94.93 \\ \cline{2-7} 
\rowcolor[HTML]{FFFFFF} 
\cellcolor[HTML]{FFFFFF} & \cellcolor[HTML]{FFFFFF} & \cellcolor[HTML]{FFFFFF}Standard & 93.36 & 94.88 & \cellcolor[HTML]{C0C0C0}96.2 & 95.65 \\
\rowcolor[HTML]{FFFFFF} 
\cellcolor[HTML]{FFFFFF} & \cellcolor[HTML]{FFFFFF} & \cellcolor[HTML]{FFFFFF}Quantization-aware & 85.23 & 85.74 & \cellcolor[HTML]{C0C0C0}87.47 & 86.31 \\
\rowcolor[HTML]{FFFFFF} 
\cellcolor[HTML]{FFFFFF} & \cellcolor[HTML]{FFFFFF} & \cellcolor[HTML]{FFFFFF}Adversarial & 93.79 & 94.96 & \cellcolor[HTML]{C0C0C0}96.25 & 95.64 \\
\rowcolor[HTML]{FFFFFF} 
\cellcolor[HTML]{FFFFFF} & \cellcolor[HTML]{FFFFFF} & \cellcolor[HTML]{FFFFFF}Mixup & 88.59 & 89.98 & \cellcolor[HTML]{C0C0C0}93.15 & 91.85 \\
\rowcolor[HTML]{FFFFFF} 
\multirow{-10}{*}{\cellcolor[HTML]{FFFFFF}\textbf{CIFAR10}} & \multirow{-5}{*}{\cellcolor[HTML]{FFFFFF}NiN} & \cellcolor[HTML]{FFFFFF}Average & 90.24 & 91.39 & \cellcolor[HTML]{C0C0C0}93.27 & 92.36 \\ \hline
\rowcolor[HTML]{C0C0C0} 
\cellcolor[HTML]{FFFFFF} & \cellcolor[HTML]{FFFFFF} & \cellcolor[HTML]{FFFFFF}Standard & 100 & 100 & 100 & 100 \\
\rowcolor[HTML]{FFFFFF} 
\cellcolor[HTML]{FFFFFF} & \cellcolor[HTML]{FFFFFF} & Adversarial & \cellcolor[HTML]{C0C0C0}100 & 83.33 & \cellcolor[HTML]{C0C0C0}100 & \cellcolor[HTML]{C0C0C0}100 \\
\rowcolor[HTML]{C0C0C0} 
\cellcolor[HTML]{FFFFFF} & \cellcolor[HTML]{FFFFFF} & \cellcolor[HTML]{FFFFFF}Mixup & 100 & 100 & 100 & 100 \\
\rowcolor[HTML]{FFFFFF} 
\cellcolor[HTML]{FFFFFF} & \multirow{-4}{*}{\cellcolor[HTML]{FFFFFF}LSTM} & Average & \cellcolor[HTML]{C0C0C0}100 & 94.44 & \cellcolor[HTML]{C0C0C0}100 & \cellcolor[HTML]{C0C0C0}100 \\ \cline{2-7} 
\rowcolor[HTML]{C0C0C0} 
\cellcolor[HTML]{FFFFFF} & \cellcolor[HTML]{FFFFFF} & \cellcolor[HTML]{FFFFFF}Standard & 100 & 100 & 100 & 100 \\
\cellcolor[HTML]{FFFFFF} & \cellcolor[HTML]{FFFFFF} & \cellcolor[HTML]{FFFFFF}Adversarial & \cellcolor[HTML]{C0C0C0}100 & \cellcolor[HTML]{FFFFFF}50.00 & \cellcolor[HTML]{C0C0C0}100 & \cellcolor[HTML]{C0C0C0}100 \\
\rowcolor[HTML]{C0C0C0} 
\cellcolor[HTML]{FFFFFF} & \cellcolor[HTML]{FFFFFF} & \cellcolor[HTML]{FFFFFF}Mixup & 100 & 100 & 100 & 100 \\
\rowcolor[HTML]{FFFFFF} 
\multirow{-8}{*}{\cellcolor[HTML]{FFFFFF}\textbf{IMDb}} & \multirow{-4}{*}{\cellcolor[HTML]{FFFFFF}GRU} & Average & \cellcolor[HTML]{C0C0C0}100 & 83.33 & \cellcolor[HTML]{C0C0C0}100 & \cellcolor[HTML]{C0C0C0}100 \\ \hline
\rowcolor[HTML]{FFFFFF} 
\cellcolor[HTML]{FFFFFF} & \cellcolor[HTML]{FFFFFF} & \cellcolor[HTML]{FFFFFF}Standard & 78.67 & \cellcolor[HTML]{C0C0C0}85.00 & 85.60 & 85.83 \\
\rowcolor[HTML]{FFFFFF} 
\cellcolor[HTML]{FFFFFF} & \cellcolor[HTML]{FFFFFF} & \cellcolor[HTML]{FFFFFF}Quantization-aware & 75.64 & 75.73 & \cellcolor[HTML]{C0C0C0}76.51 & 76.18 \\
\rowcolor[HTML]{FFFFFF} 
\cellcolor[HTML]{FFFFFF} & \cellcolor[HTML]{FFFFFF} & \cellcolor[HTML]{FFFFFF}Adversarial & 61.71 & 62.04 & \cellcolor[HTML]{C0C0C0}63.35 & 62.28 \\
\rowcolor[HTML]{FFFFFF} 
\cellcolor[HTML]{FFFFFF} & \cellcolor[HTML]{FFFFFF} & Mixup & \cellcolor[HTML]{C0C0C0}82.57 & 79.98 & 82.46 & 80.98 \\
\rowcolor[HTML]{FFFFFF} 
\cellcolor[HTML]{FFFFFF} & \multirow{-5}{*}{\cellcolor[HTML]{FFFFFF}Densenet} & Average & 74.65 & 75.90 & \cellcolor[HTML]{C0C0C0}76.98 & 76.32 \\ \cline{2-7} 
\rowcolor[HTML]{FFFFFF} 
\cellcolor[HTML]{FFFFFF} & \cellcolor[HTML]{FFFFFF} & \cellcolor[HTML]{FFFFFF}Standard & 87.00 & 93.41 & \cellcolor[HTML]{C0C0C0}95.95 & 94.44 \\
\rowcolor[HTML]{FFFFFF} 
\cellcolor[HTML]{FFFFFF} & \cellcolor[HTML]{FFFFFF} & \cellcolor[HTML]{FFFFFF}Quantization-aware & 89.71 & 91.60 & \cellcolor[HTML]{C0C0C0}97.36 & 94.79 \\
\rowcolor[HTML]{FFFFFF} 
\cellcolor[HTML]{FFFFFF} & \cellcolor[HTML]{FFFFFF} & \cellcolor[HTML]{FFFFFF}Adversarial & 88.54 & 85.81 & \cellcolor[HTML]{C0C0C0}96.77 & 89.88 \\
\rowcolor[HTML]{FFFFFF} 
\cellcolor[HTML]{FFFFFF} & \cellcolor[HTML]{FFFFFF} & Mixup & 87.73 & 89.68 & \cellcolor[HTML]{C0C0C0}96.23 & 93.10 \\
\rowcolor[HTML]{FFFFFF} 
\multirow{-10}{*}{\cellcolor[HTML]{FFFFFF}\textbf{iWildCam}} & \multirow{-5}{*}{\cellcolor[HTML]{FFFFFF}Resnet50} & Average & 88.25 & 90.13 & \cellcolor[HTML]{C0C0C0}96.58 & 93.05 \\ \hline
\end{tabular}
}
\end{table}

Table \ref{tab:rq3} presents the AUC-ROC scores of the classifiers. Regardless of the dataset, DNN, and training strategy, in most cases (27 out of 30), the classifiers trained by $Margin$ score have greater AUC-ROC scores than other classifiers, which means that the disagreement inputs and normal inputs have a bigger difference based on the $Margin$ score. Specifically, in 23 (out of 30) cases, the classifiers trained using $Margin$ score as the training data have greater than 90\% AUC-ROC scores, which indicates the classifiers are useful to distinguish the normal inputs and disagreements. Besides, most IMDb classifiers have 100\% AUC-ROC scores, the perfect results could come from the limited number of disagreements but can still prove the output-based uncertainty score is a promising indicator to represent the property of disagreements.

Figure \ref{fig:rq3_diff} also shows a few disagreements where the model has high confidence. We call them \textbf{extreme disagreements}. We utilize the $Margin$ score to set the threshold and analyze how many extreme disagreements exist and where do they come from. Concretely, we define the disagreements with $Margin$ > 0.95 as extreme. We observe that there are 3, 226, 0, and 9 extreme disagreements in MNIST, CIFAR-10, IMDb, and iWildsCam, respectively. Interestingly, all the extreme disagreements come from the disagreements between TensorFlowLite-8bit quantized model and the original model, which means this quantization moves the decision boundary a lot in some \textit{areas}. A deeper analysis could be an interesting research direction. \finding{\textbf{Finding 9}: Extreme disagreements where the original model has high prediction confidences only come from TensorFlowLite-8bit quantized models.}


\noindent\colorbox{gray!20}{\framebox{\parbox{0.96\linewidth}{
\textbf{Answer to RQ3}: Most disagreements have closer top-1 and top-2 output probabilities (i.e., smaller $Margin$ score) than normal inputs. Compared to Entropy, Gini and Least Confidence, Margin is a better metric to distinguish disagreements and normal inputs.}}}

\subsection{RQ4: Effectiveness of Retraining}
\label{subsec:rq4}
In RQ3, we observe that the disagreements are data where the model has low confidence in the prediction. We investigate if model retraining, an efficient method to improve confidence, can ensure a stable compressed model during quantization.

Table \ref{tab:rq4} presents the number of disagreements from the ID test data before and after model retraining. In most cases (18 out of 26 cases that have disagreements before retraining), the number of disagreements decreases after model retraining. However, surprisingly, there are some exceptions that the disagreements increase. For example, in \emph{MNIST, LeNet1, CoreML-8}, 6 more disagreements appear after retraining. \finding{\textbf{Finding 10}: Retraining the model using disagreements cannot always remove the disagreements.}
 
\begin{table}[h]
\caption{Number of disagreements before and after model retraining. In total: disagreements in a test dataset regardless of the quantization technique. Values in brackets are the difference. Stubborn: disagreements cannot be removed by retraining. New: disagreements appearing after retraining.}
\label{tab:rq4}
\centering
\resizebox{.85\columnwidth}{!}{
\begin{tabular}{lcccc}
\hline
 & {\color[HTML]{000000} \textbf{Before}} & {\color[HTML]{000000} \textbf{After}} & {\color[HTML]{000000} \textbf{Before}} & {\color[HTML]{000000} \textbf{After}} \\ \hline
{\color[HTML]{000000} \textbf{MNIST}} & \multicolumn{2}{c}{\textbf{LeNet1}} & \multicolumn{2}{c}{\textbf{LeNet5}} \\ \hline
\textbf{TensorFlowLite-8} & 14 & 7(-7) & 4 & 3(-1) \\
\textbf{TensorFlowLite-16} & 9 & 4(-5) & 3 & 1(-2) \\
\textbf{CoreML-8} & 2 & 8(+6) & 1 & 1(0) \\
\textbf{CoreML-16} & 0 & 0(0) & 0 & 0(0) \\
\textbf{In total} & 15 & 16(+1) & 6 & 4(-2) \\
\textbf{Stubborn} & \multicolumn{2}{c}{1} & \multicolumn{2}{c}{0} \\
\textbf{New} & \multicolumn{2}{c}{15} & \multicolumn{2}{c}{4} \\ \hline
{\color[HTML]{000000} \textbf{CIFAR-10}} & \multicolumn{2}{c}{\textbf{NiN}} & \multicolumn{2}{c}{\textbf{ResNet20}} \\ \hline
\textbf{TensorFlowLite-8} & 514 & 371(-143) & 456 & 439(-17) \\
\textbf{TensorFlowLite-16} & 24 & 26(+2) & 54 & 49(-5) \\
\textbf{CoreML-8} & 45 & 31(-14) & 181 & 56(-125) \\
\textbf{CoreML-16} & 7 & 4(-3) & 7 & 13(+6) \\
\textbf{In Total} & 540 & 401(-139) & 536 & 480(-56) \\
\textbf{Stubborn} & \multicolumn{2}{c}{47} & \multicolumn{2}{c}{100} \\
\textbf{New} & \multicolumn{2}{c}{354} & \multicolumn{2}{c}{380} \\ \hline
{\color[HTML]{000000} \textbf{IMDb}} & \multicolumn{2}{c}{\textbf{LSTM}} & \multicolumn{2}{c}{\textbf{GRU}} \\ \hline
\textbf{TensorFlowLite-16} & 8 & 7(-1) & 3 & 1(-2) \\
\textbf{CoreML-8} & 6 & 2(-4) & 2 & 0(-2) \\
\textbf{CoreML-16} & 0 & 0(0) & 0 & 0(0) \\
\textbf{In Total} & 13 & 8(-5) & 5 & 1(-4) \\
\textbf{Stubborn} & \multicolumn{2}{c}{0} & \multicolumn{2}{c}{0} \\
\textbf{New} & \multicolumn{2}{c}{8} & \multicolumn{2}{c}{1} \\ \hline
{\color[HTML]{000000} \textbf{iWildCam}} & \multicolumn{2}{c}{\textbf{DenseNet}} & \multicolumn{2}{c}{\textbf{ResNet50}} \\ \hline
\textbf{TensorFlowLite-8} & 2830 & 3319(+489) & 326 & 373(+17) \\
\textbf{TensorFlowLite-16} & 167 & 12(-155) & 187 & 143(-44) \\
\textbf{CoreML-8} & 2035 & 7(-2028) & 226 & 101(-125) \\
\textbf{CoreML-16} & 34 & 2(-32) & 16 & 27(+11) \\
\textbf{In Total} & 3834 & 3324 (-510) & 469 & 462 (-7) \\
\textbf{Stubborn} & \multicolumn{2}{c}{2230} & \multicolumn{2}{c}{24} \\
\textbf{New} & \multicolumn{2}{c}{1094} & \multicolumn{2}{c}{438} \\ \hline
\end{tabular}
}
\end{table}


\begin{figure}[!ht]
    \centering
    \subfigure[MNIST-LeNet1]{
    \includegraphics[scale=0.25]{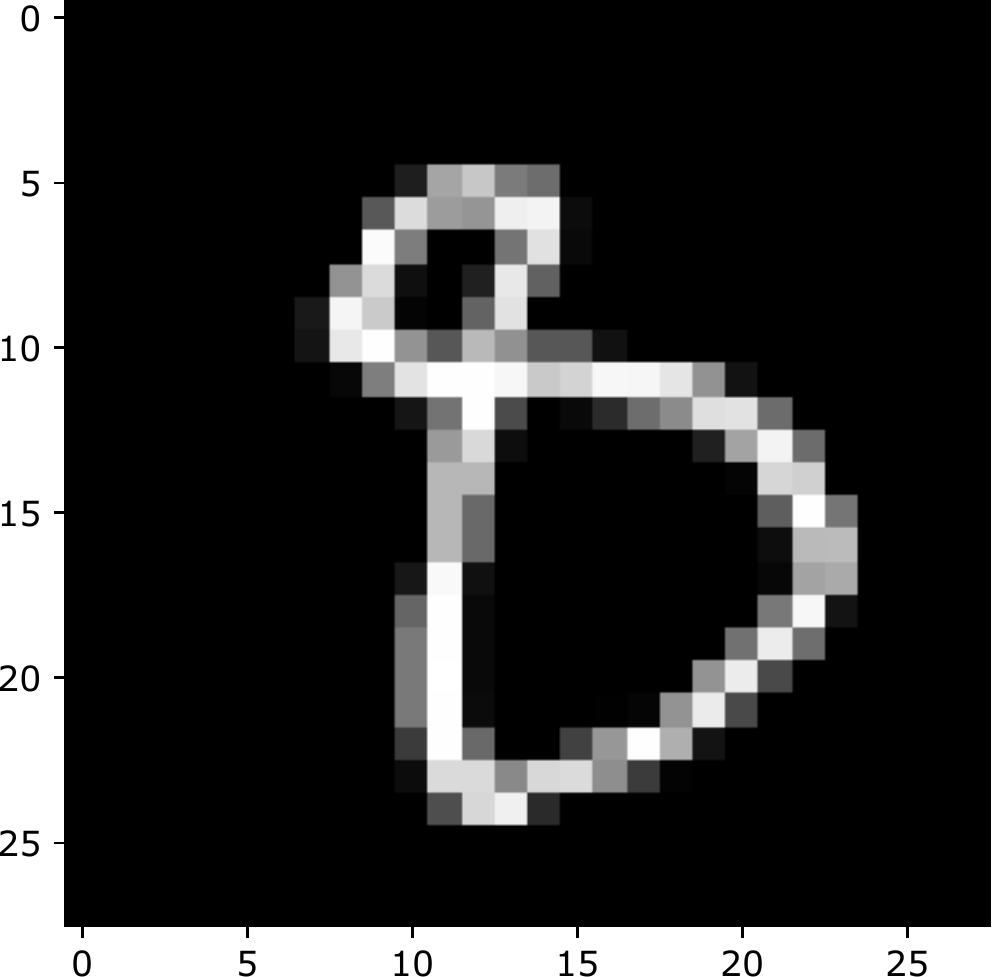}%
    }
    \subfigure[CIFAR-10-ResNet20]{
    \includegraphics[scale=0.25]{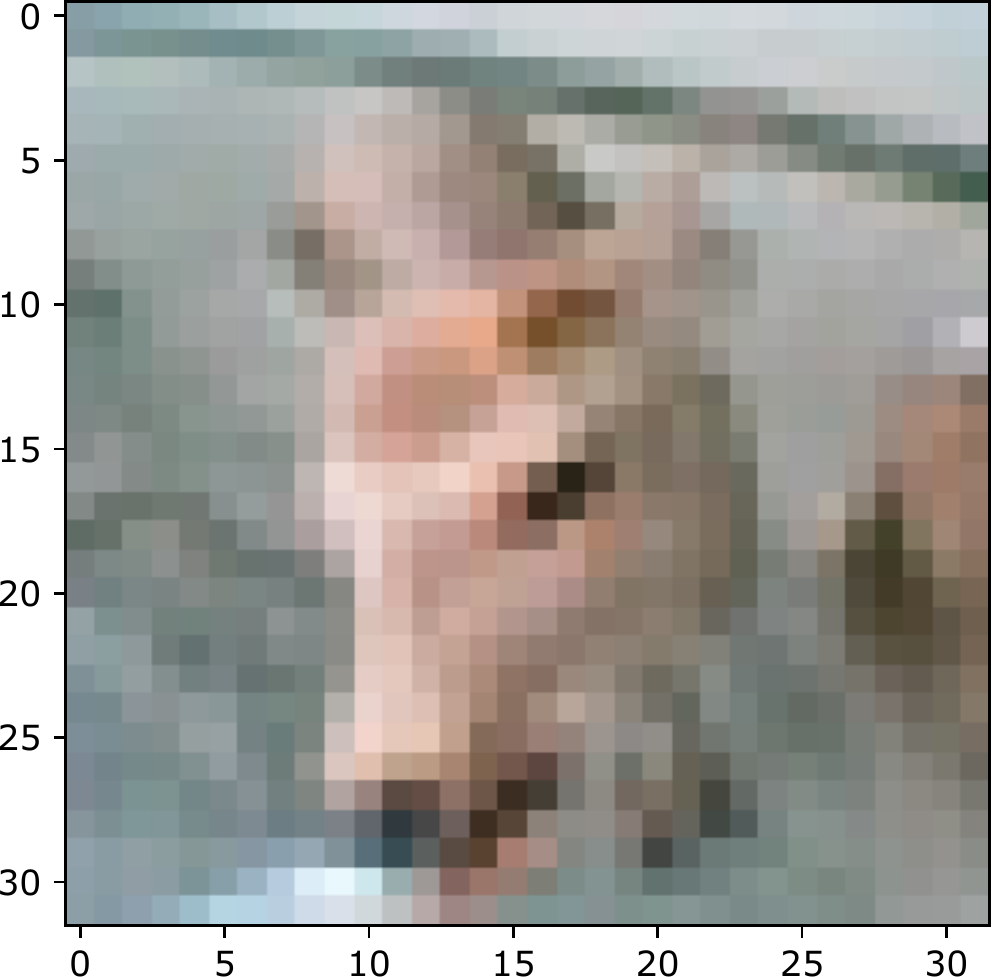}%
    }
    \caption{Examples of two stubborn disagreements. MNIST: predicted label before retraining: 1, 100\% confidence, after: 8, 100\% confidence. CIFAR-10: prediction before retraining: cat, 59\% confidence, after: deer, 92\% confidence.}
    \label{fig:rq4_example}
\end{figure}

In addition, we study whether the disagreements are really removed by model retraining. To this end, we compare if the disagreements maintain the same after retraining. For simplicity, we define the \textbf{stubborn disagreement} as the disagreement appearing both before and after retraining, and \textbf{new disagreement} as the disagreement introduced by retraining. Figure \ref{fig:rq4_example} gives two examples of stubborn disagreements. For the MNIST image, the model predicts the digital number as 0 or 9, while the true label is 8. For the CIFAR-10 image, the model hesitates to predict the animal to be a cat before retraining, and raises the confidence of this wrong prediction after retraining, while the true label is deer. Besides, we observe that the average $Margin$ score of all the stubborn disagreements before and after retraining are 0.40 and 0.56, respectively. That means although models become more confident with these stubborn disagreements after retraining, their uncertainty is still high. In Table \ref{tab:rq4}, regardless of the quantization technique, only a few stubborn disagreements remain after retraining. For example, in \emph{CIFAR-10, NiN}, only 47 (of 540) disagreements are left. However, model retraining introduces new disagreements which have the same size as without retraining. For example, in \emph{iWildCam, ResNet50}, through retraining, only 24 stubborn disagreements are left and all the other 445 are efficiently removed, but meanwhile, 438 new disagreements appear. \finding{\textbf{Finding 11}: Through model retraining, only a few stubborn disagreements remain but a similar size of new disagreements are introduced.}


\noindent\colorbox{gray!20}{\framebox{\parbox{0.96\linewidth}{
\textbf{Answer to RQ4}: Retraining fails to reduce the total number of disagreements. Though it manages to remove some existing disagreements, it introduces as many new ones.}}}

\section{Discussion}
\label{sec: discussion}

\subsection{Quantized Model Repair}
\label{subsec:remove}
We have verified that model retraining, the most common strategy to enhance performance, has limited functionality in removing disagreements. How to solve this issue is still an open problem. Based on our investigation, the disagreements are mainly the data with small $Margin$ scores by quantized models. Therefore, the main challenge is how to improve the confidence of the data. We provide two potential solutions. 1) \textbf{Online monitoring.} Before quantization, training multiple models to perform prediction can also improve confidence \cite{pmlr-v119-bielik20a}. Concretely, we can divide data into different groups based on their $Margin$ scores. For each group of data, a model is trained and quantized. 2) \textbf{Offline repair.} After quantization, building an ensemble model to perform prediction instead of the quantized model. Ensemble learning \cite{ensemble2018,LI20143120} has been proved to effectively improve the predictive performance of a single model by taking weighted average confidence from multiple models. However, both solutions will increase the storage size since more models are required. As a result, there is a trade-off between fewer disagreements and efficient model quantization. Thus, designing a robust quantization method is still an ongoing and important direction. 

\subsection{Threats to Validity}
\label{subsec:threats}
First, the threats to validity come from the selected datasets and models. Regarding the datasets, we consider both image and text classification tasks and include OOD benchmark datasets with both synthetic and natural distribution shifts. All the datasets are widely used in previous studies. As for the models, we cover two types of DNN architectures, feed-forward neural network, e.g., ResNet, and recurrent neural network, e.g., LSTM. In addition, we take into account the model complexity and apply both simple and complex ones, such as LeNet1 and ResNet50. For each dataset, we employ two different models to eliminate the influence of selected models. An interesting research direction is to repeat our experiments on other tasks, such as the regression task.

Second, the training strategies and uncertainty metrics could be other threats to validity. For the training strategies, among all possible choices, we include the four most representative and common ones. Standard training is the most basic training procedure and should be taken as the baseline. Quantization-aware training is specifically designed for quantization. Mixup training is the first and basic data augmentation approach to improve the generalization of DNNs over different distribution shifts. Adversarial training is one of the most effective techniques to promote model robustness/generalization. For the uncertainty metrics, we tend to select metrics that require as few configurations as possible. The four metrics included in this work are all solely based on the output probabilities. This is to avoid the impact of uncontrollable factors. For example, the dropout-based uncertainty metric \cite{gal2016dropout} needs to consider where to put the dropout layer and the dropout ratio.

\section{Related Work}
\label{sec: related}
\subsection{Deep Learning Testing}
\label{subsec:dlt}
As a critical phase in the software development life cycle \cite{ma2018secure}, deep learning testing ensures the functionality of DL-based systems during deployment. Multiple testing methods have been proposed in recent years \cite{zhang2020machine, kim2019guiding, tian2018deeptest, gao2020fuzz, hu2019deepmutation++}. For example, from the perspective of deep learning models, Pei \emph{et al.} proposed DeepXplore which borrows the idea from code coverage and defines neuron coverage to measure if the test set is enough or not. Later on, DeepGauge \cite{ma2018deepgauge} defines some new coverage metrics, e.g., k-multisection Neuron Coverage and Neuron Boundary Coverage, and demonstrates their effectiveness compared to the basic neuron coverage. From the perspective of test data, several test generation \cite{xie2019deephunter, guo2018dlfuzz, riccio2021deepmetis, dola2021distribution} and test selection \cite{feng2020deepgini, chen2020practical, li2019boosting} approaches have been proposed. Gao \emph{et al.} proposed SENSEI \cite{gao2020fuzz} which utilizes genetic search to find the best image transformation methods (e.g., image rotate) to generate the suitable data for training a more robust model. Chen \emph{et al.} proposed PACE \cite{chen2020practical} which uses clustering methods and MMD-critic algorithm to select a small size of test data to estimate the accuracy of the model. However, all of these works test the model before quantization, while our study mainly focuses on the analysis of the difference between the models before and after quantization.

There are two studies closely related to our work \cite{xie2019diffchaser, tian2021fast}. Both of them generate test inputs that have a different output between the original and compressed models. However, these works did not 1) study the properties of such disagreements; 2) try to solve the disagreements; 3) consider natural distribution shift, all of which are considered in our work.

\subsection{Empirical Study for Deep Learning Systems}
\label{subsec:empirical}
Empirical software engineering is one general way to practical analyze software systems. In recent years, multiple empirical studies for deep learning systems have been conducted to help understand such complex systems. 

The empirical study by Zhang \emph{et al.} \cite{Zhang2019AnES} pointed out that model migration is one of the top-three common programming issues in developing deep learning applications. Noticing the lack of benchmark understanding of the migration and quantization, Guo \emph{et al.} \cite{guo2019empirical} investigated, for deployment process, the performance of trained models when migrated/quantized to real mobile and web browsers. They focus on the impacts of the deployment process on prediction accuracy, time cost, and memory consumption. In addition to the accuracy, we further evaluate the robustness of a model, especially considering the synthetic and natural distribution shifts in the test data. Chen \emph{et al.} \cite{Chen2021AnES} studied the faults when deploying deep learning models on mobile devices. Especially, they apply TensorFlowLite and CoreML in the deployment, which is also considered in our study. The difference with our study is that their empirical study explores the failures related to data preparation (datatype error), memory issue, dependency resolution error, and so on, while our study focuses on the differential behavior during deployment and retraining. Hu \emph{et al.} \cite{hu2021towards} verified that model quantization has opposite impacts over different tasks in the setting of active learning. For example, after quantization, the model is less accurate in the image classification task while exhibiting better performance in the text classification task. In our study, since the labels of all data are available, we apply standard training instead of active learning. 

\section{Conclusion}
\label{sec: conclusion}
In this paper, we conducted a systematically study to characterize and help people understand the behaviors of quantized models under different data distributions. Our results reveal that there are more disagreement inputs in data with distribution shift than in the original test data. Quantization-aware training is a useful training strategy to produce a model that has fewer disagreements after quantization. The disagreements are those data that have high uncertainty scores, and the $Margin$ score is a more effective indicator to distinguish the normal inputs and disagreements. More importantly, we also demonstrated that the commonly used approach -- retraining the model with disagreements has limited usefulness to remove the disagreements and repair quantized models. Based on our findings, we provide two future research directions to solve the disagreement issue. To support further research, we released our code, models (before and after quantization) to be a new benchmark for studying the quantization problem.

\bibliographystyle{ACM-Reference-Format}

\bibliography{main}

\end{document}